# One Proxy Device Is Enough for Hardware-Aware Neural Architecture Search

BINGQIAN LU, University of California, Riverside, United States
JIANYI YANG, University of California, Riverside, United States
WEIWEN JIANG, George Mason University, United States
YIYU SHI, University of Notre Dame, United States
SHAOLEI REN*, University of California, Riverside, United States

Convolutional neural networks (CNNs) are used in numerous real-world applications such as vision-based autonomous driving and video content analysis. To run CNN inference on various target devices, hardware-aware neural architecture search (NAS) is crucial. A key requirement of efficient hardware-aware NAS is the fast evaluation of inference latencies in order to rank different architectures. While building a latency predictor for each target device has been commonly used in state of the art, this is a very time-consuming process, lacking scalability in the presence of extremely diverse devices. In this work, we address the scalability challenge by exploiting *latency monotonicity* — the architecture latency rankings on different devices are often correlated. When strong latency monotonicity exists, we can re-use architectures searched for one proxy device on new target devices, without losing optimality. In the absence of strong latency monotonicity, we propose an efficient proxy adaptation technique to significantly boost the latency monotonicity. Finally, we validate our approach and conduct experiments with devices of different platforms on multiple mainstream search spaces, including MobileNet-V2, MobileNet-V3, NAS-Bench-201, ProxylessNAS and FBNet. Our results highlight that, by using just one proxy device, we can find almost the same Pareto-optimal architectures as the existing per-device NAS, while avoiding the prohibitive cost of building a latency predictor for each device.

## 1 INTRODUCTION

Convolutional neural networks (CNNs) are a most commonly used class of deep neural networks, offering human-level inference accuracy for numerous real-world applications such as vision-based autonomous driving and video content analysis [21]. Going beyond the contentional server-only platforms, CNNs have been deployed on increasingly diverse devices and platforms, including mobile, ASIC and edge devices [46]. As the foundation of a CNN, the neural architecture can greatly affect the resulting model performance such as accuracy and latency. Thus, optimizing the architecture through hardware-aware neural architecture search (NAS) is crucial and being actively studied [5, 13, 34, 40, 41, 45].

The exponentially large search space consisting of billions of or even more architectures renders NAS a very challenging task [15, 40, 41, 43, 45, 47]. The key reason is that evaluating and ranking the architectures in terms of metrics of interest (e.g., accuracy and latency) can be extremely

*Corresponding author: Shaolei Ren (sren@ece.ucr.edu)

Authors' addresses: Bingqian Lu, blu029@ucr.edu, University of California, Riverside, 900 University Ave., Riverside, California, 92521, United States; Jianyi Yang, jyang239@ucr.edu, University of California, Riverside, 900 University Ave., Riverside, California, 92521, United States; Weiwen Jiang, wjiang8@gmu.edu, George Mason University, 4400 University Drive, Fairfax, Virginia, 22030, United States; Yiyu Shi, yshi4@nd.edu, University of Notre Dame, 257 Fitzpatrick Hall, Notre Dame, Indiana, 46556, United States; Shaolei Ren, sren@ece.ucr.edu, University of California, Riverside, 900 University Ave., Riverside, California, 92521, United States.

time-consuming. As a result, many studies have been focused on reducing the cost[1] of training and evaluating the architecture accuracy, including reinforcement learning-based NAS with accuracy evaluated based on a small proxy dataset [52], differentiable NAS [45], one-shot or few-shot NAS [4, 9, 51], NAS assisted with an accuracy predictor [15, 43], among many others.

In addition to speeding up accuracy evaluation, reducing the cost of assessing the inference latency on a target device is equally important for efficient hardware-aware NAS [9, 19, 33, 40]. The naive method of measuring the latency for each architecture can lead to a total search time exceeding several weeks or even months, whereas using the floating-point operations (FLOPs) as a device-agnostic proxy may not accurately reflect the true inference latency on different devices [40]. As a result, state-of-the-art (SOTA) hardware-aware NAS has mainly relied on device-specific latency lookup tables or predictors [5, 10, 13, 15, 19, 34, 43, 47].

Nonetheless, building a latency predictor for a target device requires significant engineering efforts and can be very slow. For example, [10] measures average inference latencies for 5k sample DNNs on a mobile device and uses the results to build a latency lookup table for that specific device. Assuming the ideal scenario of 20 seconds for each measurement (to average out randomness per the TensorFlow guideline [22]) and non-stop measurement, it can take 27+ hours to build the latency predictor for one single device [10]. Similarly, it is reported by [15] that 350k records are collected for building a latency predictor for just one device. Even by measuring latencies on six devices in parallel, the authors of [29] report on OpenReview that they spent *one month* to collect latency data on the small NAS-Bench-201 space and build latency predictors for another two datasets on the FBNet space. More recently, kernel-level latency predictors that capture complex processing flows of different neural execution units are proposed, but it takes up to 4.4 days for just collecting the latency measurements on one edge device [50]. All these facts highlight the crucial point that building a latency predictor for a target device — a key step of SOTA hardware-aware NAS — is costly and cannot be taken for granted as free lunch.

Worse yet, the target devices for CNN deployment are extremely diverse, ranging from mobile CPUs, ASIC, edge devices to GPUs. For example, even for the mobile devices alone, as shown in Fig. 1, there are more than two thousand system-on-chips (SoCs) available in the market, and only top 30 SoCs can each have over 1% of the share [46]. Importantly, the diverse set of devices have different latency collection pipelines, programming environment, and/or hardware domain knowledge requirement [29]. Thus, in the presence of extremely diverse target devices, the combined cost of building device-specific latency predictors for hardware-aware NAS is prohibitively high and increasingly becoming a key bottleneck for scalable hardware-aware NAS. In addition, this challenge is further magnified by the fact that building device-specific latency predictors is *not* a one-time cost: varying the input resolution and/or output classes also requires new latency predictors (e.g., two device-specific latency predictors are built, each for one dataset, on the FBNet space [29]). Consequently, how to efficiently scale up hardware-aware NAS for extremely diverse target devices has arisen as a critical challenge.

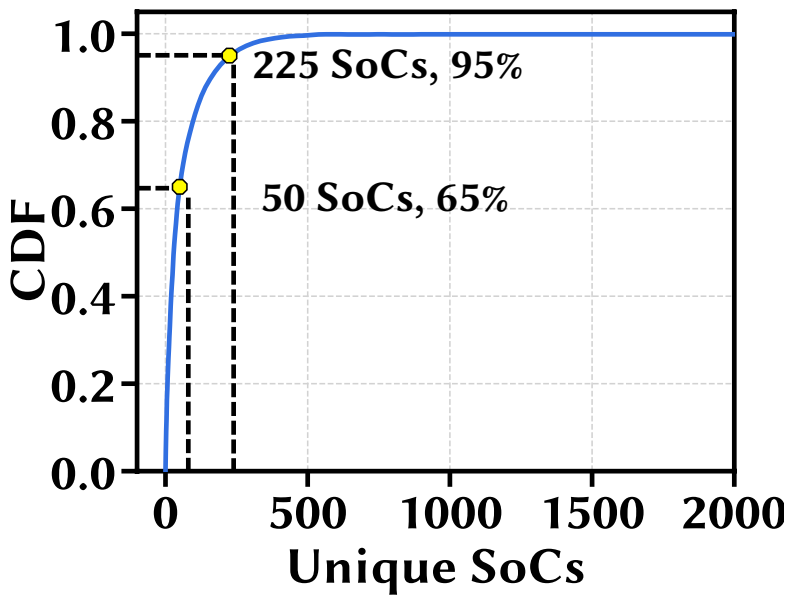

Fig. 1. Device statistics for Facebook users as of 2018 [46].

**Contributions.** In this paper, we focus on reducing the total latency evaluation cost for scalable hardware-aware NAS in the presence of diverse target devices across different platforms (e.g., mobile platform, FPGA platform, desktop/server GPU, etc.). Concretely, we show that latency monotonicity commonly exists among different devices, especially devices of the same platform. Informally, latency monotonicity means that the ranking orders of different architectures' latencies are correlated on two or more devices. Thus, with latency monotonicity, building a latency predictor for just one device that serves as a proxy — rather than for each individual target device as in state of the art [9, 15, 29] — is enough. Even when a target device has a weak monotonicity with the default proxy device (e.g., a mobile phone proxy vs. a target edge TPU), we use an efficient adaptation technique which, by measuring latencies of a small number of architectures on the target device, significantly boosts the latency monotonicity between the adapted proxy device and the target device.

[1]In this paper, "cost" also interchangeably refers to computational complexity: a higher complexity requires more computational resources (measured in, e.g., machine hours) and hence a higher monetary cost, too.

We validate our approach by considering various search spaces and running experiments with devices of different platforms, including mobile, desktop GPU, desktop CPU, edge devices and FPGA. Our results show that, using just one proxy device, there is almost no Pareto optimality loss compared to architectures specifically optimized for each target device. In addition, we also consider the recent latency datasets [19, 29, 50], and confirm further that one proxy device is enough for hardware-aware NAS.

## 2 STATE OF THE ART AND LIMITATIONS OF HARDWARE-AWARE NAS

In this section, we provide an overview of the existing (hardware-aware) NAS algorithms as well as SOTA approaches to reducing the performance evaluation cost, and highlight their limitations.

### 2.1 Overview

Neural architecture is a key design hyperparameter that affects the inference accuracy and latency of DNN models. In Fig. 2, we show an example architecture, which is found by searching over the possible layer-wise kernel sizes, expansion ratio, and block depth in the MobileNet-V2 search space using evolutionary search [9].

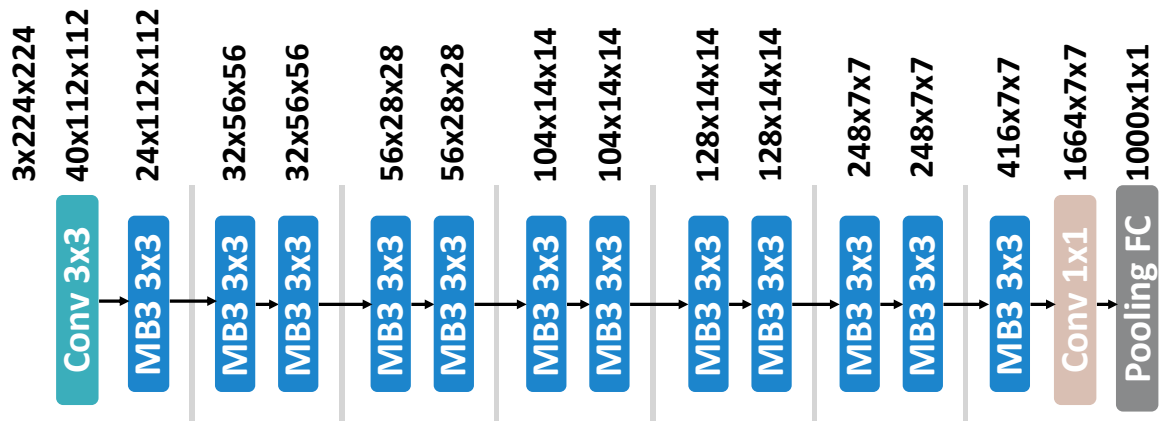

Fig. 2. An example architecture in the MobileNet-V2 search space, which achieves 70.2% accuracy on ImageNet and 71ms average inference latency on S5e. The text "$Z_1$ x $Z_2$ x $Z_3$" the input size for each layer.

The available architecture space is exponentially large, often consisting of billions of or even more choices (e.g., >$10^{19}$ in [9]). To address the complexity challenge, NAS has recently been proposed to efficiently automate the discovery of neural architectures that exceed the performance of expert-designed architectures [52]. Next, we provide a summary of existing NAS algorithms.

*2.1.1 NAS Without a Supernet.* Many prior NAS algorithms can be broadly viewed as "NAS *without* a supernet", where the search process is entangled with the model training process [35, 40, 52]. Specifically, as illustrated in the left subfigure of Fig. 3, the NAS process is governed by a controller (e.g., a reinforcement learning agent): given each candidate architecture produced by the controller, the model is trained on the training dataset and then evaluated for its performance, based on which the controller produces another candidate architecture. This process repeats until convergence or the maximum search iteration is reached. Techniques to reduce the search cost include training on part of the training dataset, a small proxy dataset, using Bayesian optimization or reinforcement learning to reduce the number of sampled candidate architectures, parameterizing the architectures and using gradients of the loss to guide the search and training simultaneously, among others [31, 34, 38–40, 52]. Nonetheless, the search cost for even a single device can still take up to 100+ GPU hours, lacking scalability in the presence of numerous heterogeneous devices [10, 45].

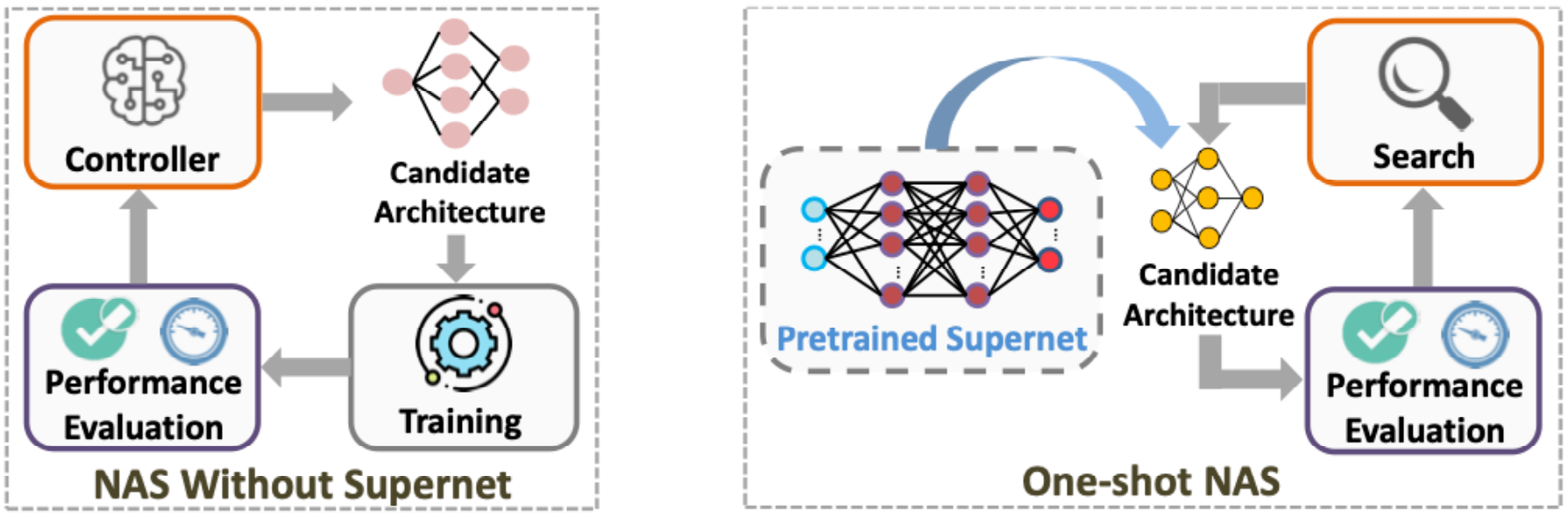

Fig. 3. Overview of NAS algorithms. Left: NAS without a supernet. Right: One-shot NAS with a supernet.

*2.1.2 One-shot NAS.* In view of the extremely diverse devices and platforms for model deployment, one-shot NAS and its variants such as few-shot NAS have recently been proposed to reduce the search cost by exploiting the weight sharing mechanism [4, 5, 9, 14, 23, 36, 48, 51]. Concretely, as illustrated in the right subfigure of Fig. 3, the key idea of one-shot NAS is to decouple the training process from the search process: pre-train a super large model (called *supernet*) whose weight is shared among all the candidate architectures, and then use a separate search process to discover optimal architectures that inherit the weights from the supernet. For example, in SOTA algorithms such as APQ, ChamNet, BigNAS and FBNet-V3 [9, 14, 15, 43, 48], a supernet is pre-trained first, which is then followed by a search process based on evolutionary algorithms or reinforcement learning to find an optimal architecture.

While pre-training the supernet is more costly than training an individual network, the training cost is one-time[2] for each learning task and, when amortized over hundreds of target devices, will be much more affordable. For example, with the recent once-for-all algorithm [9], the amortized training cost for each target device is around 12 hours given a modest size of 100 devices, and further less given more devices.

## 2.2 Current Practice for Reducing the Cost of Performance Evaluation

With a $O(1)$ model training cost incurred by one-shot NAS, the cost of performance evaluation — accuracy and latency evaluation — increasingly becomes a bottleneck.

**Accuracy evaluation.** For each candidate architecture, the time needed to evaluate the inference accuracy (even on a small proxy/validation dataset) is in the order of minutes. Thus, to expedite the accuracy evaluation, SOTA NAS algorithms have leveraged an accuracy predictor: first measuring the accuracies of sample architectures (extracted from the supernet) and then building a machine learning model [14, 15, 43]. Therefore, the candidate architectures can be ranked based on their predicted accuracies, speeding up the runtime process for NAS. Since the inference accuracy is evaluated based on the testing dataset, the accuracy predictor is device-independent and can be re-used for different target devices, incurring a fixed one-time cost of $O(1)$.

**Latency evaluation.** Measuring the actual latency for each candidate architecture takes about 20 seconds or more (to average out the random variations as per TensorFlow-Lite guideline [22] and also suggested by [10]). Meanwhile, the total number of candidate architectures sampled by a NAS algorithm is typically in the order of 10k or even more [12, 15, 40], thus settling the total latency evaluation time to be 50+ hours for just *one* target device.

[2]With the optimal architecture found by NAS, additional model updates (e.g., by training over the entire dataset or fine-tuning the weights) may still be needed to further improve the accuracy, but this will typically not affect the accuracy rankings of different architectures [34] and is orthogonal to NAS whose goal is to decide an optimal neural architecture.

Table 1. Cost Comparison of Hardware-aware NAS Algorithms for $n$ Target Devices.

| Algorithm | Search Method | Model Training | Accuracy Evaluation | Latency Evaluation | Total Cost (Machine-hours) |
|---|---|---|---|---|---|
| MNasNet [40] | RL | $O(n)$ | $O(n)$ | $O(n)$ | $6912n$ |
| FBNet [45] | Gradient | $O(n)$ | $O(n)$ | $O(n)$ | $216n$ |
| ProxylessNAS [10] | Gradient | $O(n)$ | $O(n)$ | $O(n)$ | $(200 + c_L)n$ |
| NetAdapt [47] | Loop | $O(1)$ | $O(n)$ | $O(n)$ | $c_T + (c_A + c_L)n$ |
| APQ [43] | Evolutionary | $O(1)$ | $O(1)$ | $O(n)$ | $2400 + c_A + c_L n$ |
| ChamNet [15] | Evolutionary | $O(1)$ | $O(1)$ | $O(n)$ | $c_T + c_A + c_L n$ |
| Once-for-All [9] | Evolutionary | $O(1)$ | $O(1)$ | $O(n)$ | $1200 + c_A + c_L n$ |

Using the FLOPs as a device-agnostic proxy cannot accurately reflect the true latency rankings of different architectures on a target device [40]. Instead, to reduce the latency evaluation cost, SOTA hardware-aware NAS algorithms have most commonly used latency predictors — profiling/measuring the latencies for sample architectures in advance and then building a latency predictor (either a lookup table or machine learning model) [9, 10, 15, 50]. Then, the latency predictor is utilized to guide the NAS process, without measuring the actual latency on the target device.

### 2.3 Limitations

Despite the recent progress, SOTA hardware-aware NAS algorithms still cannot scale up in view of the extremely diverse target devices for model deployment.

**Summary of total search cost.** Given $n$ target devices, we summarize in Table 1 the total search costs, measured in machine-hours, of a few representative hardware-aware NAS algorithms. If the quantitative evaluation cost is not reported for an algorithm, we use $c_T, c_A, c_L$ to denote its model training cost, accuracy evaluation cost, and latency evaluation cost, respectively. Empirically, for each device, $c_L$ is in the order of at least a few tens of hours [10, 19] or even hundreds of hours [29, 50]. Thus, we can see that the latency evaluation cost is a significant or even dominant part of the total search cost, especially when $n$ increases.

While the actual execution time of NAS may be further reduced by parallel processing, the total cost in terms of machine-hours does not decrease. For example, latency measurements on multiple devices in parallel and assigning more GPUs for supernet training can both speed up the overall NAS process, but the total resources needed by NAS still remain unchanged (or possibly even higher due to communications overheads among GPUs for distributed training). For this reason, machine-hour is a more accurate and widely-used metric for total resource expenditure in NAS [9, 40, 43, 45].

**Challenges.** In the current practice, building a latency predictor for each target device requires significant engineering efforts and can be very slow, while it is often excluded from the total cost calculation [15, 30, 42, 43, 45, 47]. Moreover, the diverse set of target devices have different latency collection pipelines, programming environment, and/or hardware domain knowledge requirement, all of which add to the significant challenges of building a latency predictor [29].

The challenges of building latency predictors have been increasingly recognized and motivated some latest studies on latency predictors to facilitate hardware-aware NAS research. For example, [29] releases latency datasets/predictors for six devices on the NAS-Bench-201 space and FBNet space. Even by measuring latencies in parallel, the authors of [29] report on OpenReview that they spent *one month* to collect latency measurement. Another recent study [50] builds a kernel-level latency predictor, taking up 1–4.4 days for latency measurement on each device depending on how powerful the device is. Nonetheless, these approaches are not scalable, and the latency predictors built by these studies are all specific to their limited set of devices.

We can conclude that, in the presence of extremely diverse target devices, the combined cost of building latency predictors for hardware-aware NAS is prohibitively high at $O(n)$. This has increasingly become a bottleneck for scalability.

## 3 PROBLEM FORMULATION, INSIGHTS, AND PRACTICAL CONSIDERATION

We present the problem formulation for hardware-aware NAS, show the key insights for when we can reduce the latency evaluation cost to $O(1)$, and finally discuss practical considerations.

### 3.1 Problem Formulation

The general problem of hardware-aware NAS can be formulated as follows:

$$\max_{\mathbf{x}\in\mathcal{X}} \max_{\omega_{\mathbf{x}}} \; accuracy(\mathbf{x}, \omega_{\mathbf{x}}) \tag{1}$$

$$s.t., \quad latency(\mathbf{x}; \mathbf{d}) \leq \overline{L}_{\mathbf{d}} \tag{2}$$

where $\mathbf{x}$ represents the architecture, $\mathcal{X}$ is the search space under consideration, $\omega_{\mathbf{x}}$ is the network weight given architecture $\mathbf{x}$, $\overline{L}_{\mathbf{d}}$ is the average inference latency constraint, and $\mathbf{d} \in \mathcal{D}$ denotes a device with $\mathcal{D}$ being the device set. Note that $accuracy(\mathbf{x}, \omega_{\mathbf{x}})$ is measured on a dataset independent of the device $\mathbf{d}$, and can also be replaced with a certain loss function (e.g., cross entropy). By varying $\overline{L}_{\mathbf{d}}$ between its feasible range $[\overline{L}_{\mathbf{d},\min}, \overline{L}_{\mathbf{d},\max}]$, we can obtain a set of *Pareto*-optimal architectures, denoted by $\mathcal{P}_{\mathbf{d}} = \{\mathbf{x}^*(\overline{L}_{\mathbf{d}}; \mathbf{d}), \text{ for } \overline{L}_{\mathbf{d}} \in [\overline{L}_{\mathbf{d},\min}, \overline{L}_{\mathbf{d},\max}]\}$.

**Remark.** We offer the following remarks on the problem formulation. First, due to the non-convexity and combinatorial nature, the obtained architectures by using approximate methods (e.g., evolutionary search [43]) to solve Eqns. (1)(2) may not be globally Pareto-optimal in a strict sense; instead, the notation of Pareto-optimality (or simply, optimality) in the context of NAS usually means a satisfactory architecture that outperforms or is very close to SOTA results [1, 14, 40]. Second, as recently shown in [29], the inference latency and energy of an architecture on a device are very strongly correlated. That is, an energy constraint can be implicitly mapped to a corresponding latency constraint. Thus, like in [5, 13, 40, 45, 47], we only consider the inference latency constraint in our formulation for the convenience of presentation.

### 3.2 Key Insights

By observing the NAS problem in Eqns. (1)(2), achieving $O(1)$ latency evaluation cost may seem very unlikely. The reason is that the inference latency $latency(\mathbf{x}; \mathbf{d})$ is highly device-specific — with a new device, the latency function will change in general, and so will the Pareto-optimal architectures accordingly. We notice, however, that the Pareto-optimal architectures for two different devices can actually be identical if their latency functions are *monotonic*, as formally defined and proved below.

Definition 1 (Latency Monotonicity). *Given two different devices $\mathbf{d}_1 \in \mathcal{D}$ and $\mathbf{d}_2 \in \mathcal{D}$, if $latency(\mathbf{x}_1; \mathbf{d}_1) \geq latency(\mathbf{x}_2; \mathbf{d}_1)$ and $latency(\mathbf{x}_1; \mathbf{d}_2) \geq latency(\mathbf{x}_2; \mathbf{d}_2)$ hold simultaneously for any two neural architectures $\mathbf{x}_1 \in \mathcal{X}$ and $\mathbf{x}_2 \in \mathcal{X}$, then the two devices $\mathbf{d}_1$ and $\mathbf{d}_2$ are said to satisfy latency monotonicity.* ■

Proposition 3.1. *If two devices $\mathbf{d}_1 \in \mathcal{D}$ and $\mathbf{d}_2 \in \mathcal{D}$ strictly satisfy latency monotonicity, then they have the same set of Pareto-optimal architectures, i.e., $\mathcal{P}_{\mathbf{d}_1} = \mathcal{P}_{\mathbf{d}_2}$, where $\mathcal{P}_{\mathbf{d}_i} = \{\mathbf{x}^*(\overline{L}_{\mathbf{d}_i}; \mathbf{d}_i), \text{ for } \overline{L}_{\mathbf{d}_i} \in [\overline{L}_{\mathbf{d}_i,\min}, \overline{L}_{\mathbf{d}_i,\max}]\}$ for $i = 1, 2$.*

Proof. Define $\mathcal{X}_{\overline{L}_{\mathbf{d}_1},\mathbf{d}_1}$ as the set of architectures satisfying $latency(\mathbf{x};\mathbf{d}_1) \leq \overline{L}_{\mathbf{d}_1}$. By latency monotonicity, we can find another constraint $\overline{L}_{\mathbf{d}_2}$ such that $\mathcal{X}_{\overline{L}_{\mathbf{d}_1},\mathbf{d}_1} = \mathcal{X}_{\overline{L}_{\mathbf{d}_2},\mathbf{d}_2}$. In other words, the latency constraint $latency(\mathbf{x};\mathbf{d}_1) \leq \overline{L}_{\mathbf{d}_1}$ is equivalent to $latency(\mathbf{x};\mathbf{d}_2) \leq \overline{L}_{\mathbf{d}_2}$. Therefore, device-aware NAS formulated in Eqns. (1)(2) for devices $\mathbf{d}_1$ and $\mathbf{d}_2$ are equivalent, sharing the same set of Pareto-optimal architectures. □

Proposition 3.1 guarantees that, for any two devices satisfying latency monotonicity, we only need to run device-aware NAS on *one* device, avoiding the cost of numerous latency measurements and building a separate latency predictor for each device. The key reason is that in NAS, it is the architecture's accuracy and latency performance ranking that really matters for Pareto-optimality. Consequently, if latency monotonicity is satisfied among all the target devices, the latency evaluation cost can be kept as $O(1)$.

### 3.3 Practical Consideration

To quantify the degree of latency monotonicity in practice, we use the metric of Spearman's Rank Correlation Coefficient (**SRCC**), which lies between -1 and 1 and assesses statistical dependence between the rankings of two variables using a monotonic function. The greater the SRCC of CNN latencies on two devices, the better the latency monotonicity. SRCC of 0.9 to 1.0 is usually viewed as *strongly* dependent in terms of monotonicity [3].

While Proposition 3.1 does not strictly hold when the SRCC is less than 1.0, we note that a sufficiently high SRCC (e.g., around 0.9 in our experiments) is already good enough in practice. This is due in great part to imperfection/approximation in other aspects of the NAS process. Concretely, in SOTA hardware-aware NAS algorithms [15, 40, 43], the accuracy predictor (or the accuracy measured on a small proxy dataset) only has a SRCC value of around 0.9 with the true accuracy. Thus, given the imperfection of accuracy evaluation, strictly satisfying the latency monotonicity does not offer substantial benefits.

## 4 LATENCY MONOTONICITY IN THE REAL WORLD

We now investigate latency monotonicity in the real world and show that it commonly exists among devices, especially of the same platform.

### 4.1 Intra-Platform Latency Monotonicity

We empirically show the existence of strong latency monotonicity among devices of the same platform, including mobile, FPGA, desktop GPU and CPU.

**Mobile platform.** We first empirically measure the actual latencies of CNN models on four mobile devices: Samsung Galaxy **S5e**, **TabA**, **Lenovo** Moto Tab, and **Vankyo** MatrixPad Z1 (a low-end device). The details of deveice specifications are listed in Table 2. We randomly sample 10k models from the MobileNet-V2 space [37] (details in Section 6). Then, we deploy these models on the four devices and calculate their average inference latencies. We show the actual latencies on S5e, Lenovo, Vankyo versus TabA in Fig. 4(a), where each dot represents one CNN model.

We see that when the sampled CNN models run faster on TabA, they also become faster on the other devices. In Fig. 4(a), the maximum standard deviation (denoted by the vertical line within each bin) is 1.3% for Vankyo, while it is negligibly 0.6% and 0.84% for Lenovo and S5e. Thus, latency monotonicity is well preserved on these devices. We further show the SRCC values of these 10k sampled model latencies on our four mobile device in Fig. 4(b) with heatmap. We see that SRCC between any pair of our mobile devices is larger than 0.98, implying strong latency monotonicity.

| Device | Abbrev. | Chipset | CPU (GHz) | Cores | RAM (GB) | RAM Freq. (MHz) | Peak Perf. (GFLOPs/sec) | Mem. Bandwidth (GB/sec) |
|---|---|---|---|---|---|---|---|---|
| Samsung Galaxy Tab S5e | **S5e** | Snapdragon 670 | 2 | 8 | 4 | 1866 | 40.6 | 14.93 |
| Samsung Galaxy Tab A | **TabA** | Snapdragon 429 | 2 | 4 | 2 | 933 | 15.3 | 7.46 |
| Lenovo Moto Tab | **Lenovo** | Snapdragon 625 | 2 | 8 | 2 | 933 | 26.5 | 7.5 |
| Vankyo MatrixPad Z1 | **Vankyo** | N/A | 1.5 | 4 | 1 | 933 | N/A | N/A |

Table 2. Device specifications. Full details are not available for Vankyo.

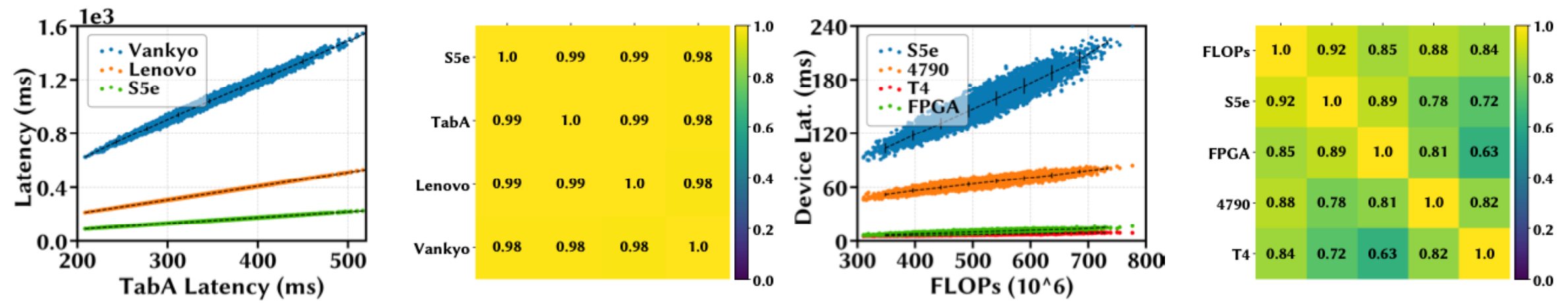

(a) (b) (c) (d)

Fig. 4. Empirical measurement of latency monotonicity. (a)(c) Black vertical lines denote the standard deviation of latency data points within each bin, with the center denoting the average. (b)(d) SRCC of 10k sampled model latencies on different pairs of devices.

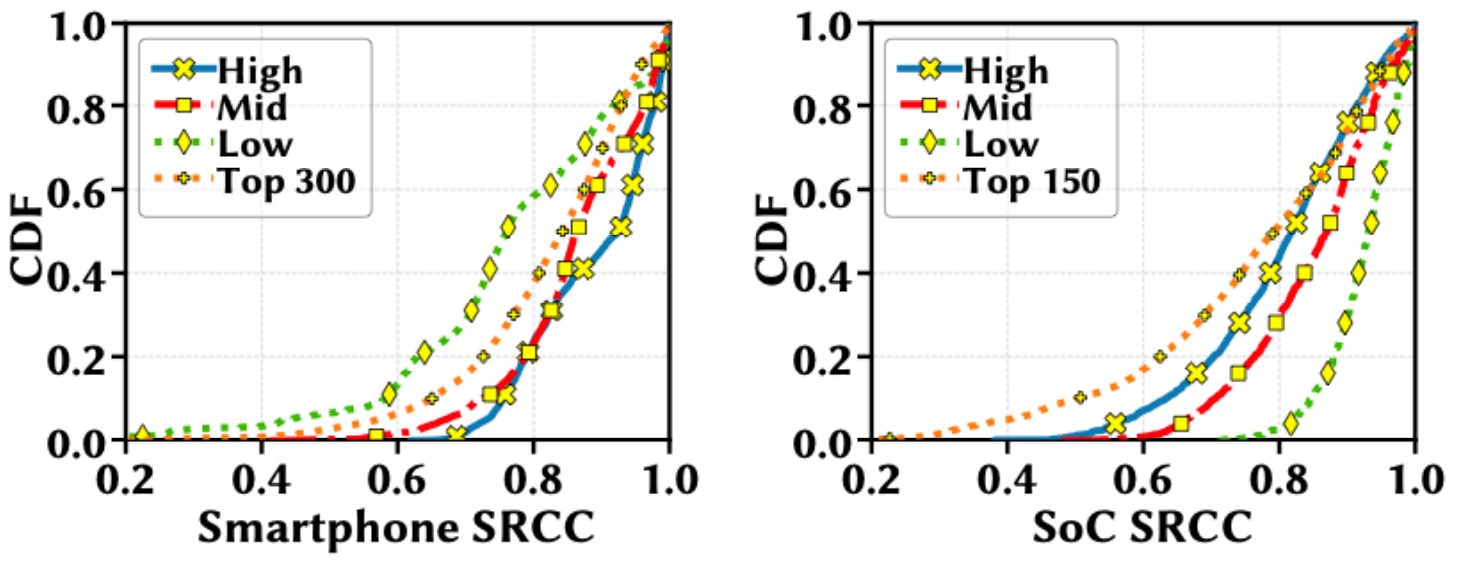

(a) Top-300 Mobile Phones (b) Top-150 Mobile Phones

Fig. 5. CDF of SRCC values of DNN models on mobile phones and SoCs. The annotation "high/mid/low" represents the highest/middle/lowest 33.3% of the devices.

**AI-Benchmark data.** To examine latency monotonicity at scale, we resort to the AI-Benchmark dataset showing DNN inference latency measurements on diverse hardware [2]. Considering top-300 smartphones (ranging from Huawei Mate 40 Pro to Sony Xperia Z3) and top-150 mobile SoCs (ranging from HiSilicon Kirin 9000 to MediaTek Helio P10) ranked by the metric "AI-Score" [25], we show in Fig. 5 the SRCC values of latency rankings based on the 22 DNN models including both floating-point and quantized models (e.g., MobileNet-V2-INT8 and MobileNet-V2-FP16) listed in the dataset. We see that latency monotonicity is well preserved at scale. For example, among the top-100 mobile phones, SRCC values among 50+% of *any* device pairs are higher than 0.9 (a very strong ranking correlation). While the AI-Benchmark dataset is built for orthogonal purposes and includes models from different search spaces, the resulting SRCC values, along with our own experiments, still provide a good reference and show reasonable latency monotonicity for mobile devices at scale.

**Other platforms.** Going beyond the mobile platform, we also perform experiments to show latency monotonicity on other platforms: desktop CPU, GPU and FPGA.

| Index | Computation Design | | | Communication Design | | |
|---|---|---|---|---|---|---|
| | $T_m$ | $T_n$ | $T_m(d)$ | $I_p$ | $O_p$ | $W_p$ |
| 1 | 160 | 12 | 576 | 11 | 8 | 13 |
| 2 | 160 | 12 | 576 | 5 | 5 | 22 |
| 3 | 160 | 12 | 576 | 12 | 13 | 17 |
| 4 | 160 | 12 | 576 | 10 | 10 | 10 |
| 5 | 130 | 12 | 832 | 10 | 10 | 10 |
| 6 | 100 | 16 | 832 | 10 | 10 | 10 |
| 7 | 220 | 8 | 704 | 10 | 10 | 10 |
| 8 | 100 | 16 | 832 | 6 | 14 | 10 |
| 9 | 100 | 18 | 704 | 10 | 10 | 10 |

Table 3. Nine FPGA specifications on Xilinx ZCU 102 board.

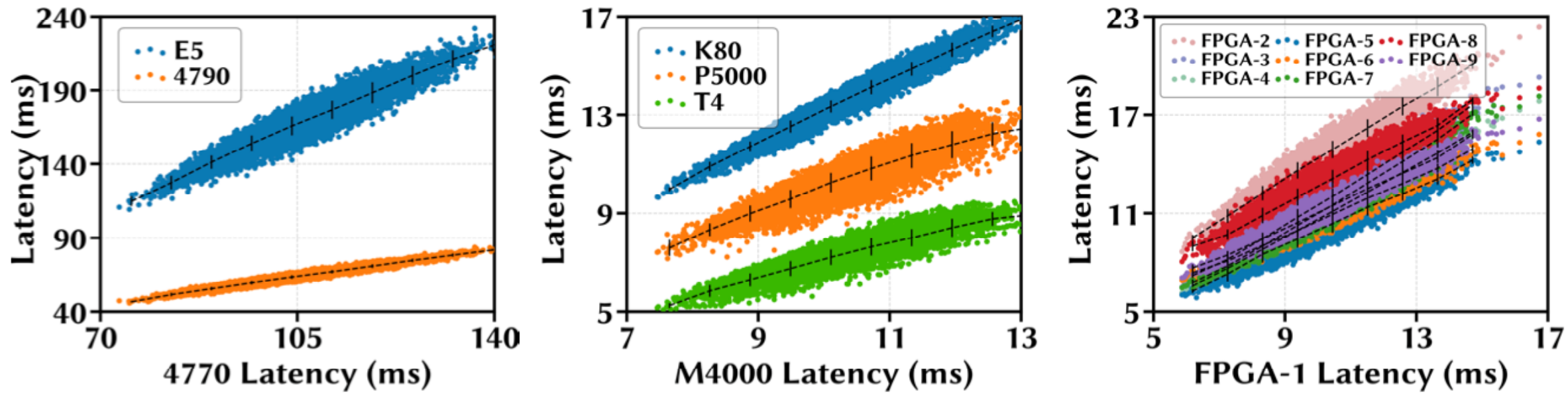

(a) Desktop CPU (b) Desktop GPU (c) FPGA

Fig. 6. Latency monotonicity on non-mobile platforms. Black vertical lines denote the standard deviation of latency data points within each bin, with the center denoting the average.

We build latency lookup tables for three desktop CPUs: Intel Core i7-**4790**, Intel Core i7-**4770** HQ, and **E5**-2673 v3. In addition, we consider four NVIDIA GPUs: Tesla **T4**, Tesla **K80**, Quadro **M4000**, and Quadro **P5000**. For the FPGA platform, we configure nine subsystems for an Xilinx ZCU 102 FPGA board to create nine different FPGAs following the hardware design space in [26]. The detailed configuration for FPGAs is shown in Table 3. "Computation Design" is the computation subsystem design, $T_m$, $T_n$ are loop tiling parameters for input and output feature maps, and $T_m(d)$ denotes the parameter for depth-wise separable convolution. "Communication Design" represents the communication subsystem design, where $I_p$, $O_p$, and $W_p$ are communication ports allocated for input feature maps, output feature maps and weights, respectively. We measure CNN model latency on nine Xilinx ZCU 102 boards shown in Table 3, using the performance model in [26].

We consider latencies for the same set of 10k models as in Fig. 4, and plot the results in Figs. 6 and 7, respectively. We see that *within* each platform, latency monotonicity is generally very well preserved, with most SRCC values close to or above 0.9+. In addition, we also show in Fig. 7 the SRCC between the model FLOPs and the actual inference latency, confirming the prior observation that FLOP may not accurately reflect the true latency performance [29, 45, 50].

Next, to complement our own measurement, we also examine latency monotonicity by leveraging third-party latency predictors and measurements results on other devices. The results are available in Appendix B.1 and further corroborate our finding.

## 4.2 Inter-Platform Latency Monotonicity

We choose one FPGA (Xilinx ZCU 102), one desktop CPU (Intel Core i7-**4790**), and one desktop GPU (Tesla **T4**) as cross-platform devices. We show the latency monotonicity results and SRCC values for the same set of 10k models in Figs. 4(c) and 4(d), respectively. It can be seen that latency rankings are only moderately correlated for cross-platform devices. The SRCC values are lower than in the case of mobile device pairs (Fig. 4(b)), since mobile devices often differ significantly from desktops/FPGAs.

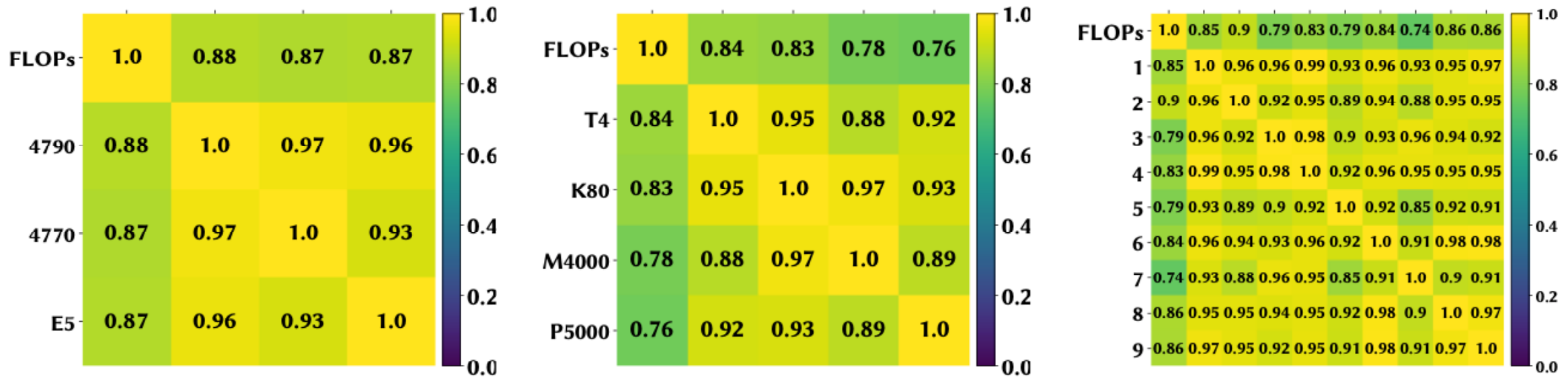

(a) Desktop CPU (b) Desktop GPU (c) FPGA

Fig. 7. SRCC of 10k sampled model latencies on different pairs of non-mobile devices. Specification of nine FPGAs in Fig. 7(c) is listed in Table 3.

Our finding is also confirmed in the appendix by considering the six cross-platform devices on the NAS-Bench-201 [17] and FBNet[45], and four devices on MobileNet-V3 using nn-Meter [50].

### 4.3 Roofline Analysis

We now explain the empirically observed latency monotonicity based on roofline analysis, which is a methodology for visual representation of hardware platform's *peak* performance as a function of the operational intensity, which identifies the bottleneck of the system [44]

Fig. 8(a) shows the theoretical roofline model of two mobile devices (Samsung Galaxy S5e and TabA) plotted according to their reported hardware specification listed in Table. 2. When operational intensity is low (linear slope region), memory bandwidth is the limiting factor for program speed (i.e., memory-bound); when operational intensity is high (horizontal region), peak FLOPs rate becomes the bottleneck (i.e., compute-bound).

Suppose that we have two devices $\mathbf{d}_1$ and $\mathbf{d}_2$ with memory bandwidths $B_{\mathbf{d}_1}$ and $B_{\mathbf{d}_1}$, respectively, and two CNN models of architectures $\mathbf{x}_1$ and $\mathbf{x}_2$ with operational intensities $OI_{\mathbf{x}_1}$ and $OI_{\mathbf{x}_2}$, respectively. Next, we show that latency monotonicity is guaranteed to hold for two devices if CNN models are either memory-bound or compute-bound on both devices.

*Memory-bound.* In the memory-bound region, the slope in the roofline model of a device is the bandwidth, and the resulting performance is the bandwidth multiplied by the program's operational intensity. Assuming that $\mathbf{x}_1$ is slower than $\mathbf{x}_2$ on device $\mathbf{d}_1$ without loss of generality, we have $\frac{FLOP_{\mathbf{x}_1}}{OI_{\mathbf{x}_1} \cdot B_{\mathbf{d}_1}} > \frac{FLOP_{\mathbf{x}_2}}{OI_{\mathbf{x}_2} \cdot B_{\mathbf{d}_1}}$. Then, by multiplying both sides by $\frac{B_{\mathbf{d}_1}}{B_{\mathbf{d}_2}}$, we obtain $\frac{FLOP_{\mathbf{x}_1}}{OI_{\mathbf{x}_1} \cdot B_{\mathbf{d}_2}} > \frac{FLOP_{\mathbf{x}_2}}{OI_{\mathbf{x}_2} \cdot B_{\mathbf{d}_2}}$, i.e., $\mathbf{x}_1$ is also slower than $\mathbf{x}_2$ on device $\mathbf{d}_2$. Thus, latency monotonicity holds for the two devices $\mathbf{d}_1$ and $\mathbf{d}_2$.

*Compute-bound.* Likewise, if CNN models fall into the compute-bound region for two devices, then we can also establish latency monotonicity using a similar logic.

For search spaces with models that span across both memory-bound and compute-bound regions, the latency monotonicity may not be strong (which we shall address in this paper). Moreover, the roofline analysis only provides a *sufficient* condition for latency monotonicity under the assumption that devices run at their peak performances (in terms of FLOPs/sec). Thus, we experimentally show the actual performance of CNN models on our four mobile devices shown in Table 2.

We measure the actual attainable peak performance of our four devices with the tool in [24], a roofline model specially for mobile SoCs. Our results show that the sampled CNN models (with 2.6 to 5.4 FLOPs/Byte) are all in the *compute*-bound region for the devices. We randomly sample 10000 models from the MobileNet-V2 [37]. The empirical roofline results are shown in Fig. 8(b). The operational intensity of the sampled models ranges from 2.6 to 5.4 FLOPs/Byte, while the devices' actual performances as shown in Fig. 9 are much lower than their peaks and vary for different models. Specifically, the ridge operational intensity of S5e, Lenovo, and Vankyo are less than or around 2 FLOPs/Byte, while TabA has a threshold of 3 FLOPs/Byte. Thus, most of our sampled models reside in the compute-bound region of these devices, except for those with operational intensity less than 3 FLOPs/Byte on TabA. This partially explains the strong latency monotonicity that we empirically observe in Fig. 4(b).

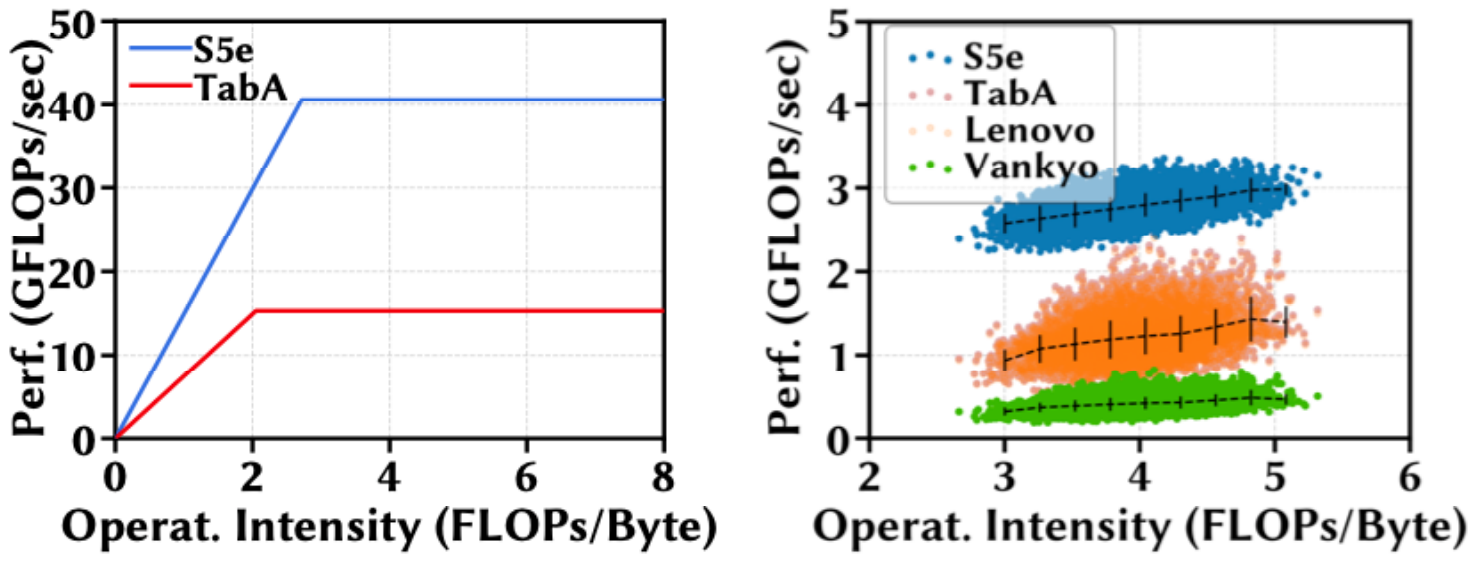

(a) Theoretical roofline model (b) Sampled models

Fig. 8. (a) Theoretical roofline model is plotted according to hardware specification of S5e and TabA. (b) Black vertical lines denote the standard deviation of data within each bin, with the center denoting the average.

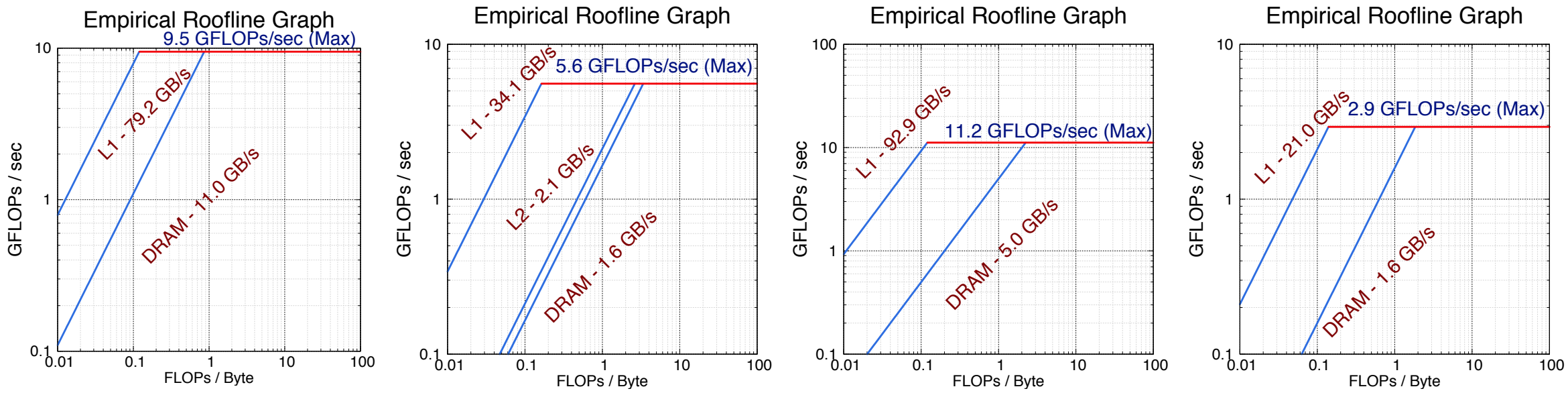

(a) S5e (b) TabA (c) Lenovo (d) Vankyo

Fig. 9. Empirical roofline models of devices in Table 2 measured with Gables [24].

## 5 HARDWARE-AWARE NAS WITH ONE PROXY DEVICE

Section 4 demonstrates good latency monotonicity among devices of the same platform, but this is not always the case, especially for devices across different platforms. To address the cases of low monotonicity, we propose efficient transfer learning based on the proxy device.

### 5.1 Necessity of Strong Latency Monotonicity

We first highlight the necessity of strong latency monotonicity for finding optimal architectures on the target device. An interesting and challenging case is when latency monotonicity is not satisfied, and this is not uncommon in practice as shown in Section 4. In such cases, the optimal architectures searched on one device can be far from optimality on another device. To see this point, we show in Fig. 11(a) the performance of architectures found on different devices using the MobileNet-V2 search space. All latencies are measured on S5e (Mobile), and the architectures directly found for

**Algorithm 1** Hardware-Aware NAS With One Proxy Device

**Inputs:** Target device $\mathbf{d}$, proxy device $\mathbf{d}_0$ with its latency predictor $L_{\mathbf{d}_0}(\mathbf{x})$ and Pareto-optimal architecture set $\mathcal{P}_{\mathbf{d}_0}$, small sample architecture set $\mathcal{A}$, SRCC threshold $S_{th}$
**Output:** Pareto-optimal architecture $\mathcal{P}_{\mathbf{d}}$
Measure $latency(\mathbf{x}; \mathbf{d})$ for $\mathbf{x} \in \mathcal{A}$;
Estimate SRCC $S_{\mathbf{d},d}$ for sample architectures in $\mathcal{A}$;
**if** $S_{\mathbf{d},\mathbf{d}_0} \geq S_{th}$ **then**
    Set $\mathcal{P}_{\mathbf{d}} = \mathcal{P}_{\mathbf{d}_0}$, or re-run NAS (e.g., evolutionary search) based on $L_{\mathbf{d}_0}(\mathbf{x})$ to obtain $\mathcal{P}_{\mathbf{d}}$;
**else**
    Use Eqn. (3) to obtain $L_{\mathbf{d}_0,\mathbf{d}}(\mathbf{x})$ based on measured $latency(\mathbf{x}; \mathbf{d})$ for $\mathbf{x} \in \mathcal{A}$;
    Run NAS based on $L_{\mathbf{d}_0,\mathbf{d}}(\mathbf{x})$ to obtain $\mathcal{P}_{\mathbf{d}}$;
**end if**
Measure latencies for architectures $\mathbf{x} \in \mathcal{P}_{\mathbf{d}}$ on device $\mathbf{d}$, and remove non-Pareto-optimal ones from $\mathcal{P}_{\mathbf{d}}$;

S5e are Pareto-optimal ones. Nonetheless, when performing NAS on two other (proxy) devices — 4790 (Desktop CPU) and T4 (Desktop GPU) — which both have low SRCC values with S5e, the searched architectures are highly sub-optimal. Thus, given weak latency monotonicity, the Pareto optimality of $\mathcal{P}_{\mathbf{d}_0}$ on the proxy device $\mathbf{d}_0$ does not hold on the target device $\mathbf{d}$, calling for remedies to boost the latency monotonicity.

## 5.2 Overview

Our scalable hardware-aware NAS approach is illustrated in Fig. 10 and described in Algorithm 1.

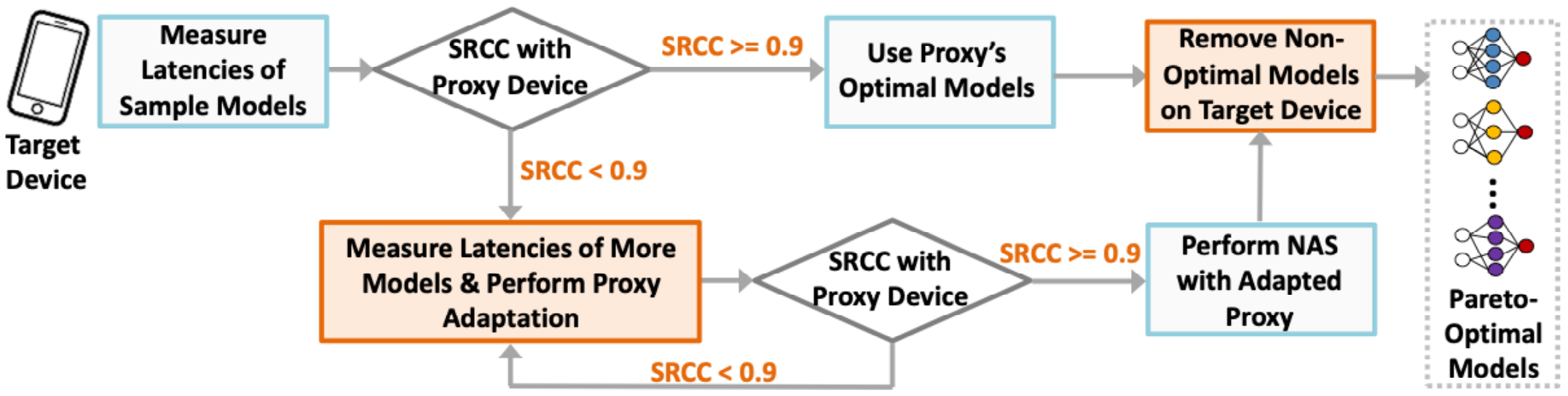

Fig. 10. Overview of using one proxy device for hardware-aware NAS.

**Prerequisite.** The prerequisite step is to select a proxy device $\mathbf{d}_0$ and run SOTA hardware-aware NAS to find a set $\mathcal{P}_{\mathbf{d}_0}$ of Pareto-optimal architectures for the proxy.

**Checking latency monotonicity.** Given a new target device, we check whether strong latency monotonicity is satisfied between the proxy device and the target device, by estimating the SRCC based on a small set of sample architectures $\mathcal{A}$ and comparing it against a threshold.

• *When strong latency monotonicity holds.* With strong latency monotonicity, the target device's Pareto-optimal architecture set $\mathcal{P}_{\mathbf{d}}$ is also likely the same as proxy device's $\mathcal{P}_{\mathbf{d}_0}$. Alternatively, we can also re-run evolutionary search based on the proxy device's latency predictor to obtain more architectures, which are in turn also likely optimal ones for the target device.

• *When strong latency monotonicity does not hold.* We propose an efficient transfer learning technique — adapting the proxy's latency predictor to the target device. By doing so, we can quickly find optimal architectures for the target device, yet *without* first measuring latencies of thousands of architectures and then building a latency predictor.

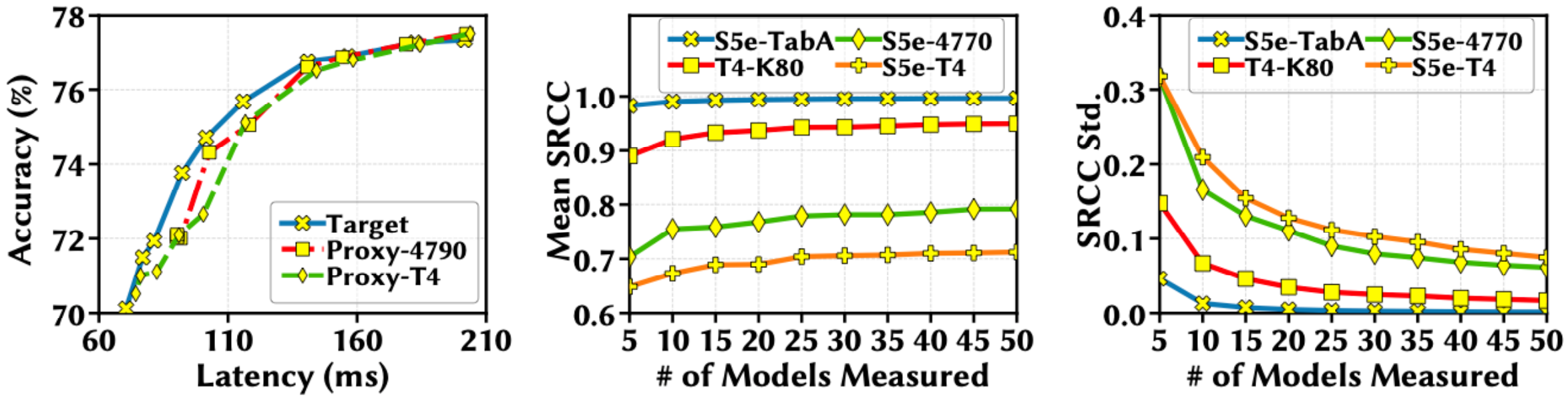

(a) NAS results with low SRCC values (b) Average Estimated SRCC (c) Standard Deviation

Fig. 11. (a) Architectures by evolutionary search in the MobileNet-V2 search space. All latencies are measured on S5e (Mobile). Architectures searched on 4790 (Desktop CPU) and T4 (Desktop GPU) are highly sub-optimal compared to those searched specifically on S5e. These two devices have SRCC of 0.78 and 0.72 with S5e, respectively. (b)(c) SRCC estimation. $X$-axis denotes the number of sample architectures we randomly select per run. We use 1000 runs to calculate the mean and standard deviation. "x-y" means the device pair is $(x, y)$.

**Removing non-Pareto-optimal architectures.** We measure the actual latencies of Pareto-optimal architectures (obtained for either the paroxy or adapted proxy device) on the target device, and remove non-Pareto-optimal architectures.

## 5.3 Prerequisite and Checking Latency Monotonicity

*5.3.1 Prerequisite.* We first select a proxy device $\mathbf{d}_0$ that preferably has good latency monotonicity with other target devices. To do so, we can first measure the latencies of a small set $\mathcal{A}$ of sample architectures (e.g., 30-50 sample architectures in our experiments) on all the target devices, and calculate the resulting SRCC values for each pair of devices based on the measured latencies. We only need to measure the overall inference latency, unlike building latency predictors which typically needs profiling the latency of each operator/layer for thousands of architectures [30, 50].

Then, we can obtain a SRCC matrix like the one shown in Fig. 4(d). Latency measurement of a small set of sample architectures is also needed to check latency monotonicity (Section 5.3.2) and hence is not an extra step. Next, we can choose a proxy device that has high SRCCs with a good number of other devices. Note that proxy device selection does not need to be very precise; instead, even though we choose a proxy device that does not have high SRCCs with many other devices, our proposed proxy adaptation technique can still significantly boost the SRCC between the selected proxy device and target devices.

For the selected proxy device, we run SOTA hardware-aware NAS to find Pareto-optimal architectures. Specifically, following the one-shot NAS approach [9, 43], we first pre-train a supernet and build an accuracy predictor. We then build a latency predictor denoted by $L_{\mathbf{d}_0}(\mathbf{x})$ based on extensive latency profiling and SOTA methods for latency prediction [19, 50]. Finally, we apply evolutionary search [15, 43], which quickly produces the Pareto-optimal architecture set $\mathcal{P}_{\mathbf{d}_0}$ by varying different latency constraints. Once the accuracy predictor and latency predictor are built, running evolutionary search takes at most a few minutes and hence is negligible.

*5.3.2 Checking latency monotonicity.* To check whether strong latency monotonicity is satisfied between the selected proxy device and a target device, we estimate the SRCC based on a small set $\mathcal{A}$ of sample architectures and then compare it against a threshold. The latency measurement for the small set of sample architectures is already performed during the proxy selection process. In Figs. 11(b) and 11(c), we can see that latency measurement based on a few sample architectures is enough to reliably estimate the SRCC value: e.g., if we set 0.9 as the SRCC threshold, then 30-50 sample architectures are sufficient. Thus, the cost for measuring latencies for the small set $\mathcal{A}$ of sample architectures is negligible compared to building a device-specific latency predictor.

### 5.4 Increasing Latency Monotonicity by Adapting the Proxy Latency Predictor

As illustrated in Fig. 11(a), in case of weak latency monotonicity, we cannot re-use the Pareto-optimal architectures found for the proxy device to a new target device. To address this issue, we propose an efficient transfer learning technique — adapting the proxy's latency predictor to the target device — to boost latency monotonicity.

*5.4.1 A close look at SOTA latency predictors.* We first review three major types of SOTA latency predictors used in hardware-aware NAS.

• **Operator-level latency predictor.** A straightforward approach is to first profile each operator [10, 15] (or each layer [6, 39]), and then sum all the operator-level latencies as the end-to-end latency of an architecture. Specifically, given $K$ operators (e.g., each with a searchable kernel size and expansion ratio), we can represent each operator using one-hot encoding: 1 means the respective operator is included in an architecture, and 0 otherwise. Thus, an architecture can be represented as $\mathbf{x} \in \{0,1\}^K \cup \{1\}$, where the additional $\{1\}$ represents the non-searchable part, e.g., fully-connected layers in CNN, of the architecture. Accordingly, the latency predictor can be written as $l = \mathbf{w}^T\mathbf{x}$, where $\mathbf{w} \in \mathbb{R}^{K+1}$ is the operator-level latency vector. This approach needs a few thousands of latency measurement samples (taking up a few tens of hours) [10, 30].

• **GCN-based latency predictor.** To better capture the graph topology of different operators, a recent study [19] uses a graph convolutionary network (GCN) to predict the inference latency for a target device. Concretely, the latency predictor can be written as $l = GCN_\Theta(\mathbf{x})$, where $\Theta$ is the learnt GCN parameter learnt and $\mathbf{x}$ is the graph-based encoding of an architecture.

• **Kernel-level latency predictor.** Another recent latency predictor is to use a random forest to estimate the latency for each execution unit (called "kernel") that captures different compilers and execution flows, and then sum up all the involved execution units as the latency of the entire architecture [50]. This approach unifies different DNN frameworks, such as TensorFlow and Onnx, into a single model graph, and hence can predict latencies for models developed using different frameworks. By encoding an architecture based on the execution units, we can also transform the latency predictor into a linear one: $l = \mathbf{w}^T\mathbf{x}$ where $\mathbf{w}$ is the vector of latencies for different execution units and $\mathbf{x}$ denotes the number of each execution unit included in an architecture. Thus, an "execution unit" in [50] is conceptually equivalent to a searchable operator in the operator-level latency predictor [10].

**Summary.** The three SOTA latency predictors use different encodings/representations for an architecture: the encoding based on searchable operators in an operator-level predictor is the simplest, while the encoding based on fine-grained execution units in a kernel-based predictor has the most details of an architecture. Despite different prediction accuracies in terms of mean squared errors, they all reflect the latency rankings on an actual device very well and hence are sufficient for serving as the proxy predictor.

*5.4.2 Adapting the proxy latency predictor.* We propose efficient transfer learning to boost the otherwise possibly weak latency monotonicity for a target device.

**Intuition.** Even though two devices have weak latency monotonicity, it does not mean that their latencies for each searchable operator are uncorrelated; instead, for most operators, their latencies can still be roughly proportional. The reason is that a more complex operator with higher FLOPs that is slower (say, 2x slower than a reference operator) on one device is generally also slower on another device, although there may be some differences in the slow-down factor (say, 2x vs. 1.9x). This is also the reason why some NAS algorithms use the device-agnostic metric of architecture FLOPs as a rough approximation of the actual inference latency [40, 41]. If we view proxy adaptation as a new learning task, this task is highly correlated with the task of building the proxy device's latency predictor, and such correlation can greatly facilitate transfer learning.

**Approach.** To explain our transfer learning approach, we consider the proxy device's latency predictor in a linear form: $L_{\mathbf{d}_0}(\mathbf{x}) = \mathbf{w}^T\mathbf{x}$, where $\mathbf{w}$ is the weight and $\mathbf{x}$ is the architecture representation (e.g., one-hot encoding of the searchable operators, penultimate layer output in a neural network-based predictor,[3] or encoding of the execution units). We measure the latencies of a small set of sample architectures $\mathbf{x} \in \mathcal{A}$ on the target device, noting that this step is also needed to check the SRCC value and incurs a negligible overhead compared to SOTA approaches (i.e., tens of hours of latency measurement [29, 50]). Then, with the latency measurement samples denoted by $(\mathbf{x}_i, y_i)$, we quickly adapt the proxy device's latency predictor as $L_{\mathbf{d}_0,\mathbf{d}}(\mathbf{x}) = \left[(\alpha\mathbf{I}^T + \mathbf{b}^T) \circ \mathbf{w}^T\right]\mathbf{x}$ tailored to the target device, by solving the following the problem:

$$\min_{\alpha,\mathbf{b}} \frac{1}{N}\sum_i \left|\left[(\alpha\mathbf{I}^T + \mathbf{b}^T) \circ \mathbf{w}^T\right]\mathbf{x} - y_i\right|^2 + \lambda|\mathbf{b}|, \tag{3}$$

where $\mathbf{I}$ is the identity vector with all the elements being 1, the operator "$\circ$" denotes the element-wise multiplication, and $\lambda \geq 0$ is a hyperparameter controlling the weight for the sparsity regularization term $|\mathbf{b}|$ and tuned based on a small validation set of architectures (20 architectures in our experiment) split from the sample architecture set $\mathcal{A}$.

The interpretation of using Eqn. (3) is as follows. First, the scaling factor $\alpha$ reflects our intuition that a more complex operator that is slower on one device is generally also slower on another device. Second, the sparsity term $\mathbf{b}$ accounts for the fact that the slow-down factors for an operator on two devices are not necessarily the same.

With $L_{\mathbf{d}_0,\mathbf{d}}(\mathbf{x})$, we essentially construct a new virtual proxy device (called adapted proxy or AdaProxy) whose latency is given by $L_{\mathbf{d}_0,\mathbf{d}}(\mathbf{x})$. Here, our goal is to increase the latency monotonicity between the new virtual proxy and the target device; we do not need to create a new latency predictor that produces accurate estimates of the absolute latency values for the target device.

If strong latency monotonicity still does not hold between AdaProxy and the target device, we can incrementally measure the latencies of another small set of sample architectures on the target device and re-solve Eqn. (3). In the majority of our experiments, 50 latency measurements on the target device are enough to achieve a strong latency monotonicity. This is negligible compared to thousands of latency profiling and measurements used by SOTA algorithms [10, 50].

Next, with the adapted latency predictor $L_{\mathbf{d}_0,\mathbf{d}}(\mathbf{x})$ that reflects the architecture latency rankings on the target device $\mathbf{d}$, we can run evolutionary search to find the set of Pareto-optimal architectures.

### 5.5 Remove non-Pareto-optimal architectures

Up to this point, we have obtained for the target device an architecture set $\mathcal{P}_\mathbf{d}$, which is the same as the proxy (or AdaProxy) device's Pareto-optimal set. While the latency monotonicity between the proxy (or AdaProxy) device and the target device is strong (e.g., SRCC around 0.9 or higher), it is not perfect. Thus, some architectures in $\mathcal{P}_\mathbf{d}$ may not be Pareto-optimal for the target device. We remove these architectures based on their actual latencies measured on the target device. Specifically, if an architecture $\mathbf{x}_1 \in \mathcal{P}_\mathbf{d}$ has a higher latency but the same or similar accuracy compared to another architecture $\mathbf{x}_2 \in \mathcal{P}_\mathbf{d}$, we can remove $\mathbf{x}_1$ from $\mathcal{P}_\mathbf{d}$.

Finally, if there is a specific latency constraint that is not satisfied by architectures in $\mathcal{P}_\mathbf{d}$, we can re-run evolutionary searches with the assistance of $L_{\mathbf{d}_0}(\mathbf{x})$, or adapted $L_{\mathbf{d}_0,\mathbf{d}}(\mathbf{x})$ if applicable, to further enlarge the set $\mathcal{P}_{\mathbf{d}}$. The key point is that we do not need to go through a very time-consuming process to build a new latency predictor specifically for the target device.

[3] If the proxy device uses a neural network-based latency predictor, we can also fix the earlier layers while updating the weights in the last few layers, instead of only updating the last single layer.

In summary, the cost for measuring latencies of a small sample set of architectures on the target device for checking latency monotonicity (and, if needed, adapting $L_{\mathbf{d}_0}(\mathbf{x})$) is negligible. Therefore, given $n$ different devices, we achieve a total latency evaluation cost of $O(1)$, which, when combined with SOTA NAS algorithms that have $O(1)$ cost for model training and accuracy evaluation [9, 15], successfully keeps the entire NAS cost at $O(1)$.

## 6 EXPERIMENT

We run experiments on multiple devices (including mobile phones, desktop GPU/CPU, ASIC, etc.) on different mainstream search spaces — MobileNet-V2, MobileNet-V3, NAS-Bench-201, and FBNet.

### 6.1 Results on MobileNet-V2

*6.1.1 Setup.* We now present the setup for our experiments on MobileNet-V2.

**Search Space.** As in [10], the backbone of our CNN architecture is MobileNet-V2 with multiplier 1.3, with the channel number in each block fixed. The search space consists of depth of each stage, kernel size of convolutional layers, and expansion ratio of each block. The depth can be chosen from "2, 3, 4", kernel size can be "3, 5, 7", and candidate expansion ratios are "3, 4, 6". There are five stages whose configurations can be searched.

**NAS Method.** We consider one-shot NAS and use the *Once-For-All* network [9] as a supernet that has the same search space as ours. We run evolutionary search to find optimal architectures for the proxy (or AdaProxy) device. Our parameter settings are: population size is 1000, parent ratio is 0.25, mutation probability is 0.1, mutation ratio is 0.25, and we search for 50 generations given each latency constraint. Evolutionary search takes less than 30 seconds for each run. To facilitate the readers' understanding, we provide a summary of evolutionary search in Appendix A, while the full details can be found in [15, 43].

**Accuracy Predictor.** The evolutionary search is assisted with by an accuracy predictor for fast architecture performance evaluation [15, 43]. Our accuracy predictor is a neural network with four fully-connected layers and updated with 176 samples on top of the predictor used in [9]. The accuracy predictor takes a 128-dimensional feature vector (which is converted from a 21-dimensional architecture configuration within the search space) as input. Fig. 12(a) compares the actual and predicted accuracies, which have a SRCC of 0.903 and root mean squared error of 1.11%. The performance of our accuracy predictor is in line with the existing NAS literature for MobileNet-based models [9]. As a result, the imperfection in the accuracy predictor explains why a strong, but not perfect, latency monotonicity (e.g., SRCC>0.9) is enough for our one-proxy approach to find Pareto-optimal architectures for a new target device.

**Latency Predictor.** We build device-specific latency predictors in the MobileNet-V2 space for our four devices listed in Table 2. Specifically, for each sample architecture, we profile the average latency of 1000 runs. We use a single thread for running the TensorFlow Lite interpreter by default. To show the accuracy of our latency predictors, we sample a few additional models and measure their actual latency on our four mobile devices. The comparison between actual and predicted latency is shown in Fig. 12, with a root mean squared error of 2.88ms on S5e, 4.69ms on TabA, 3.72ms on Lenovo, and 59.18ms on the low-end Vankyo. As corroborated by prior studies [15, 43, 47], our result shows that the predicted average latency is almost identical to the actual value.

We choose S5e mobile phone as the proxy device. Our results of using other mobile devices as the proxy are nearly the same because S5e and the other mobile devices have SRCC close to 1.0 (Fig. 4(b)), i.e., these mobile devices are almost viewed *one* device based on Proposition 3.1.

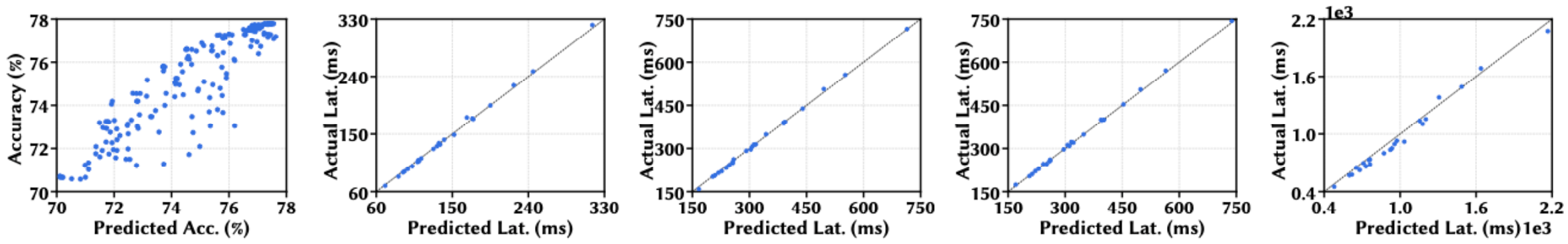

(a) Accuracy (b) S5e (c) TabA (d) Lenovo (e) Vankyo

Fig. 12. (a) Actual vs. predicted accuracy. The root mean squared error is 1.11%, and SRCC is 0.903. (b)(c)(d)(e) Measured average inference latency versus predicted latency based on latency lookup tables. The root mean squared errors for S5e, TabA, Lenovo, and Vankyo are 2.88ms, 4.69ms, 3.72ms, and 59.18ms respectively.

**Architecture Evaluation.** For a searched architecture, the actual model performance is measured. We evaluate accuracies on the ImageNet validation dataset [16], which consists of 50000 images in 1000 classes. Accuracy evaluation is run on Google Colab equipped with Tesla T4.

*6.1.2 Baselines.* We consider the following baselines for hardware-aware NAS.

**#1: Building a Latency Predictor for Each Target Device [9, 15, 19, 43].** For each device, we use the same evolutionary search described in Section 6.1.1. While the accuracy predictor is reusable across devices and evolutionary search is quick, measuring latencies of thousands of architectures to build a device-specific latency predictor (as done in the existing hardware-aware NAS [9, 19, 43]) is time-consuming. Thus, this approach has a total cost of $O(n)$ for $n$ devices [5, 15].

**#2: Heuristic Model Scaling.** There are different ways to scale a CNN to meet different latency constraints: e.g., adapt the network depth and/or width [40, 41]. Since the number of channels in our backbone network is fixed, we heuristically scale the depth of a Pareto-optimal architecture on the proxy device by increasing (for higher accuracy) or reducing (for smaller latency) the depth by up to two blocks, and transfer the scaled architecture to new target devices. This approach has $O(1)$ complexity.

The two baselines highlight that the existing hardware-aware NAS either achieves Pareto optimality but has a $O(n)$ latency evaluation cost (Baseline #1), or keeps the latency evaluation cost at $O(1)$ but loses Pareto optimality (Baseline #2). By contrast, our approach has a $O(1)$ latency evaluation cost in total, while preserving Pareto optimality.

*6.1.3 Performance of Searched Architectures.* We compare the measured top-1 accuracy on ImageNet versus average inference latency of searched architectures on each target device.

**Mobile Devices.** Fig. 13 shows the result for three different target mobile devices, all using S5e as the proxy device. The SRCC values between S5e and the target devices are all greater than or equal to 0.98 (Fig. 4(b)). We see that the architectures searched on S5e can result in almost the same $(accuracy, latency)$ tradeoff as device-specific NAS, but the additional latency evaluation cost for each target device is negligible. Further, we see that despite its $O(1)$ complexity, heuristic adaptation (baseline #2) can result in really bad architectures without performance guarantees.

**Non-Mobile Devices.** We show the results in Fig. 14 for non-mobile devices. As these devices have low SRCC values with our S5e proxy, we use Eqn. (3) to create an AdaProxy device, which has SRCC of close to 0.9 or higher with the target devices. The details of the proxy adaptation process, including the SRCC values before and after proxy adaptation, are available in Appendix B.2.

The top row shows the architectures found by evolutionary search. We see that with a low SRCC (around 0.7-0.8), the architectures searched on the proxy device are not Pareto-optimal on the target devices. With proxy adaptation, the SRCC increases significantly, and the architectures searched on the AdaProxy device are almost the same as those directly searched on the target device. This highlights the need of strong latency monotonicity between the proxy and the target device, as well as the effectiveness of our proposed proxy adaptation technique to boost the latency monotonicity. The heuristic scaling approach (Baseline #2) performs even worse than directly using the architectures searched on the proxy device, and hence are omitted.

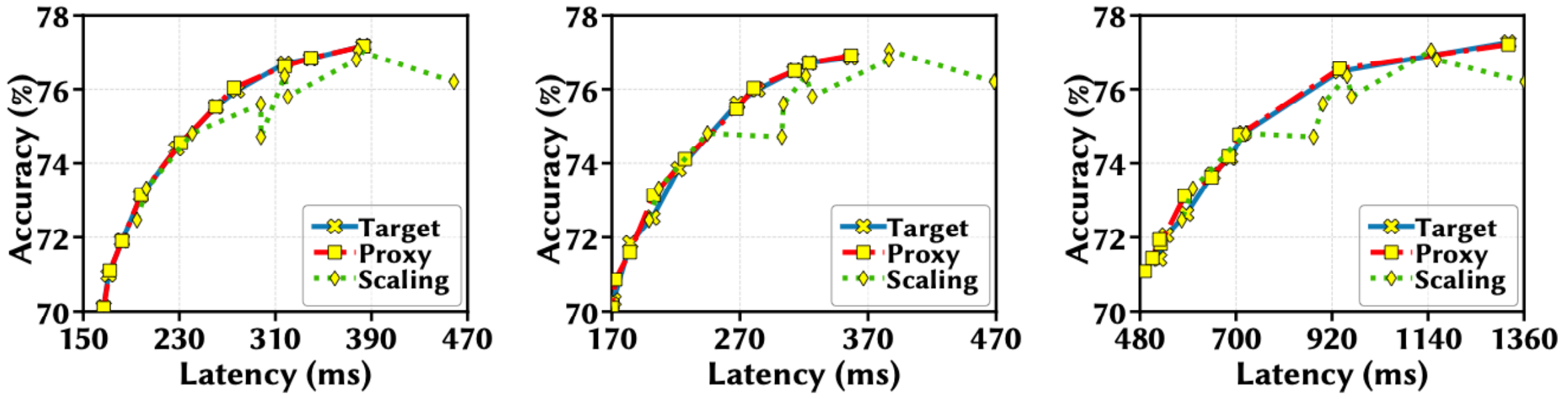

(a) TabA as target device (b) Lenovo as target device (c) Vankyo as target device

Fig. 13. Results on three different mobile target devices, using S5e as proxy device. “Target” is the baseline #1, “Proxy” means using our approach with S5e as the proxy device, and “Scaling” means heuristic scaling applied to S5e’s one Pareto-optimal architecture.

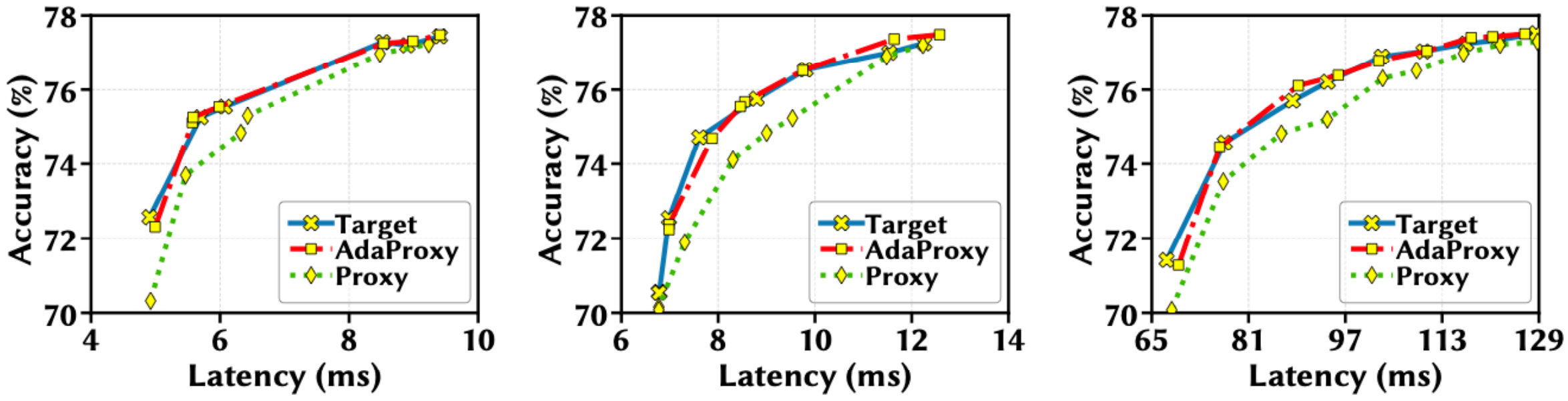

(a) Tesla T4 (b) Quadro M4000 (c) Intel i7-4770HQ

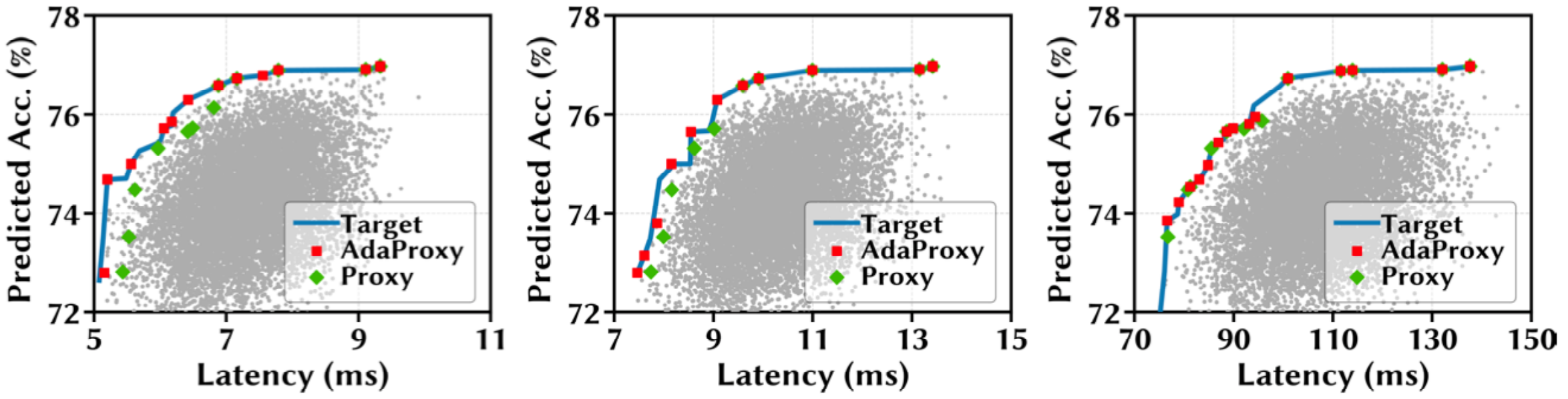

(d) Tesla T4 (e) Quadro M4000 (f) Intel i7-4770HQ

Fig. 14. Results for non-mobile target devices with the default S5e proxy and AdaProxy. The top row shows the evolutionary search results with real measured accuracies, and the bottom row shows the exhaustive search results based on 10k random architectures and predicted accuracies.

The bottom row shows exhaustive search results out of 10k randomly selected architectures, using the predicted accuracies as the true values. This is essentially considering a semi-oracle NAS process (on a small space of 10k architectures) assuming a perfect accuracy predictor. As a result, compared to evolutionary search using an imperfect accuracy predictor, it may have a more stringent requirement on the SRCC between the target device and the proxy (or AdaProxy) device. We see that, due to the low SRCC, the architectures found by using the proxy device's latency predictor may not overlap with the oracle's Pareto-optimal boundary. In fact, some of the proxy's optimal architectures can perform very poorly on the target device. For example, Fig. 14(d) shows that S5e's optimal architectures are highly sub-optimal on Tesla T4. On the other hand, with improved SRCC, the architectures found by using the AdaProxy device's latency predictor preserves Pareto optimality very well on the target devices, again demonstrating the necessity and effectiveness of our proxy adaptation technique in the presence of weak latency monotonicity between the default proxy and target device.

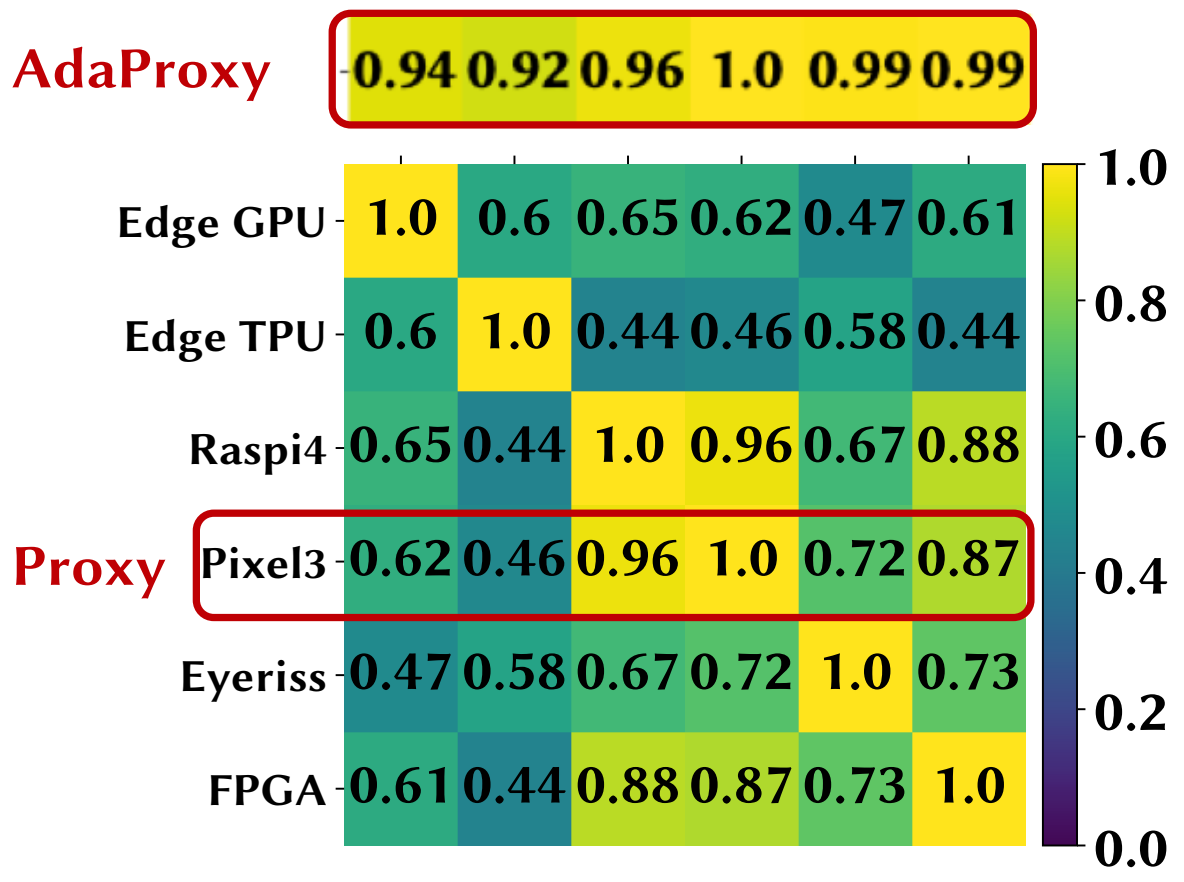

Fig. 15. SRCC for various devices in the NAS-Bench-201 search space on CIFAR-10. Pixel3 is our proxy device. SRCC values boosted with AdaProxy are highlighted.

Additional results, including settings for proxy adaptation and comparison of exhaustively searched architectures on other devices, can be found in Appendix B.2.

### 6.2 Results on NAS-Bench-201, FBNet, and nn-Meter

We now evaluate our approach on the recently released latency datasets for six different devices on NAS-Bench-201 and FBNet spaces [29], additional devices on NAS-Bench-201 [19], as well as four devices on nine different search spaces [49].

We first consider the latency results on the NAS-Bench-201 search space using the CIFAR-10 dataset [29]. Since NAS-Bench-201 represents a simple architecture space with only around 15k architectures, we consider an oracle NAS process via exhaustive search. Thus, compared to evolutionary search using an imperfect accuracy predictor, the oracle NAS process can have a more stringent requirement on the SRCC between the target device and the proxy (or AdaProxy) device. We use Pixel3 as the default proxy which, as shown in Fig. 15, does not have strong latency monotonicity with the target devices (except for Raspi4). By proxy adaptation, we can significantly boost the latency monotonicity, increasing the SRCC values to 0.9 or higher.

Next, Fig. 16 shows the optimal architectures found by using the proxy device's latency predictor, the adapted latency predictor, and the oracle, respectively. We can see that due to the pre-adaptation low SRCC values between the proxy device Pixel3 and the target devices, only a few architectures that are optimal for the proxy are still optimal for the target devices after architecture removal (Section 5.5). Moreover, even the proxy's remaining optimal architectures can be far from optimality on the target device. For example, Fig. 16(a) shows that some of Pixel3's optimal architectures deviate from the Pareto-optimal boundary on the edge GPU. By using proxy adaptation and increasing the SRCC values, the AdaProxy's optimal architectures can be efficiently transferred to target devices while preserving optimality. The proxy device Pixel3 has a high SRCC of 0.96 with Raspi4, even without proxy adaptation. Thus, as shown in Fig. 16(e), the optimality of Pixel3's architectures preserve very well on Raspi4. All these demonstrate the importance of strong monotonicity between the proxy and the target device, as well as the effectiveness of our proxy adaptation technique. for scalable hardware-aware NAS.

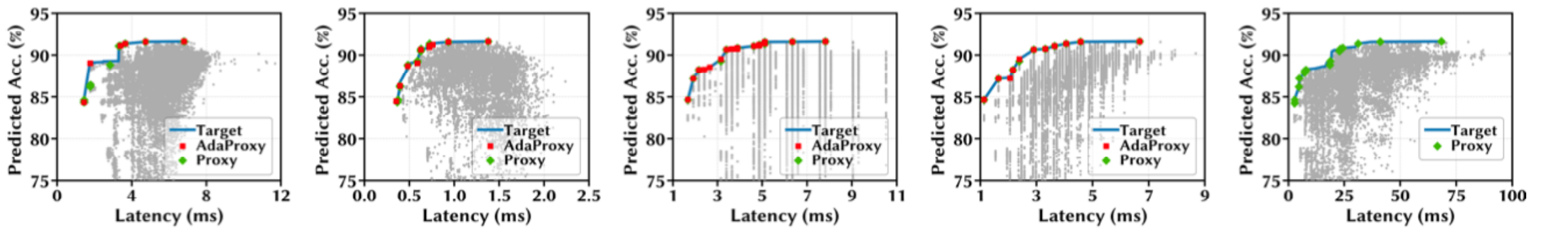

(a) Edge GPU (b) Edge TPU (c) Eyeriss (d) FPGA (e) Raspberry Pi 4

Fig. 16. Exhaustive search results for different target devices on NAS-Bench-201 architectures (CIFAR-10 dataset) [17, 29]. Pixel3 is the proxy.

Additional results, including the details of proxy adaptation and results on other search spaces, are available in Appendix B. These results further validate our approach and highlight the practical feasibility of using only one proxy device for scalable hardware-aware NAS.

## 7 RELATED WORK

The huge search space for neural architectures presents significant challenges (see [20, 30, 33, 40, 41, 45, 52] and references therein). To minimize the cost of training numerous architectures, one-shot NAS uses a super net that includes all the weights for candidate architectures [4, 5, 9, 23, 34]. In recent years, transformer-based vision algorithms have also been emerging and inspired studies transformer search to optimize the performance [18], but it is orthogonal to NAS that we focus on.

Importantly, fast evaluation of accuracy and inference latency to rank different architectures is crucial for efficient hardware-aware NAS [20, 30, 33, 40, 41, 45, 52]. To reduce the cost of accuracy evaluation, the prior studies have considered reinforcement learning with accuracy evaluated based on a small proxy dataset [52], Bayesian optimization-based NAS (to reduce the number of sampled and evaluated architectures) [35], generative approaches [27], one-shot or few-shot NAS [4, 9, 51], and NAS assisted with an accuracy predictor [15, 43]. More recently, ranking architecture accuracies based on easily-computable proxy metrics has also been studied: e.g., computing a model score based on a small minibatch of training data [1], and analyzing the neural tangent kernel (NTK) as well as the number of linear regions in the input space [11].

To expedite inference latency evaluation, the SOTA hardware-aware NAS has mainly resorted to device-specific latency predictors [5, 10, 13, 15, 19, 34, 43, 47]. Nonetheless, building even one latency predictor incurs a non-trivial upfront cost. Thus, [19, 29, 50] have recently released latency datasets and predictors, but only for a few devices due to the prohibitive time cost.

Given many diverse devices, scalability of latency evaluation is critically important. A straightforward approach is to build a meta latency predictor that incorporates hardware features as additional input [28, 32]. Nonetheless, significant drawbacks exist for this approach: (1) numerous latency measurements on a large number of heterogeneous devices are required in advance for meta-training; (2) there is a fundamental challenge for provably-good generalization to new unseen target devices that deviate significantly from the training device pool (i.e., out-of-distribution); and (3) the process of meta-learning and adaptation to new devices involves complex hyperparameter tuning, adding considerable uncertainties to the latency prediction performance. For example, in order to cover 24 devices with good generalization performance in the experiment, up to 18 heterogeneous devices are used for meta-training in which 900/4000 architecture latencies are collected for each device on the NAS-Bench-201/FBNet search space, while only the remaining 6 devices are used for testing [28]. Crucially, these meta latency predictors [28, 32] aim at producing accurate latency prediction with low prediction errors, which adds further challenges to the prediction model but is *unnecessary* for hardware-aware NAS. By contrast, what matters most is the architecture latency ranking on a target device, for which sophisticated (meta) latency predictors may not offer substantial benefits. We show both theoretically and empirically that one proxy device that has strong latency monotonicity with target devices (after proxy adaptation if needed) is enough for hardware-aware NAS, truly keeping the total latency evaluation cost at $O(1)$.

Considering a synthetic latency metric aggregated over a few devices, simultaneous multi-device NAS [13] may not meet the latency constraint or achieve Pareto optimality for any involved device. Heuristic scaling approaches, e.g., by changing the number of layers and/or channels [36, 40, 41], can limit the architecture space and hence reduce both accuracy and latency evaluation costs, but they may also miss Pareto-optimal architectures because of their coarse scaling granularity. Architecture FLOPs is a device-agnostic proxy metric, but it cannot accurately reflect the true latency ranking of architectures on real devices [15, 29, 40]. While various proxy metrics (e.g., NTK [11]) have been considered for accuracy evaluation, our approach of using one proxy device is the first to address a complementary challenge of fast latency evaluation in the presence of many diverse devices.

## 8 CONCLUSION

In this paper, we efficiently scale up hardware-aware NAS for diverse target devices. Concretely, we demonstrate latency monotonicity among different devices, and propose to use just one proxy device's latency predictor for NAS. When latency monotonicity is not satisfied between the proxy device and the target device, we propose an efficient transfer learning technique — adapting the proxy's latency predictor to the target device — to boost latency monotonicity. Overall, our approach results in a much lower total cost of latency evaluation, yet without losing Pareto optimality. For evaluation, we conduct experiments with different devices of different platforms on mainstream search spaces, including MobileNet-V2, MobileNet-V3, NAS-Bench-201 and FBNet spaces.

## ACKNOWLEDGEMENT

Bingqian Lu, Jianyi Yang, and Shaolei Ren are supported in part by the NSF under grants CNS-1910208 and CNS-2007115. Yiyu Shi is supported in part by the NSF under grants CNS-1822099 and CNS-2122220. We are grateful to the anonymous reviewers and our shepherd, Sergey Blagodurov, for their valuable comments.

# Appendix

In the appendix, we provide a summary of evolutionary search used in our experiment and additional experimental results.

## A SUMMARY OF EVOLUTIONARY SEARCH

### A.1 Description

To facilitate the readers' understanding, we provide a summary of the widely-used evolutionary search process for NAS, taking the MobileNet-V2 search space as example. More details of using evolutionary search in hardware-aware NAS can be found in [15, 43].

In our experiment, the total number of searchable blocks is 21, divided into five stages plus the last convolutional layer. Thus, we can use two 21-dimension vectors to represent the kernel size and expansion ratio of each block, respectively, and one 5-dimension vector to denote the depth of each stage. The depth can be chosen from "2, 3, 4", the kernel size can be "3, 5, 7", and the candidate expansion ratios are "3, 4, 6". Each individual member in evolutionary search consists of these three vectors. Here is an example individual: {"kernel_size": [5, 3, 5, 7, 5, 3, 5, 3, 7, 7, 5, 7, 5, 3, 3, 5, 5, 3, 5, 5, 3], "expansion_ratio": [3, 3, 4, 6, 4, 3, 4, 6, 4, 3, 6, 4, 3, 4, 3, 4, 4, 3, 3, 4, 3], "depth": [2, 2, 2, 2, 3]}.

To run evolutionary search, we first randomly sample the initial population of individuals according to the population size. Next, we evaluate the *fitness* of each individual in the population, where the fitness function is defined as:

$$(t-1) \cdot accuracy + t \cdot latency \tag{4}$$

where $t \in [0, 1]$ is the weight parameter to balance the tradeoff between accuracy and latency of each individual model, and *accuracy* and *latency* are predicted values given by the accuracy and latency predictors, respectively. By varying $t \in [0, 1]$, we can obtain a set of Pareto-optimal architectures.

For each evolutionary search iteration, we select the fittest individuals as parents for reproduction, which will survive in the next generation and also breed new individuals through crossover. For example, if our population size is 1000 and the parent ratio is 0.25, we have 250 fittest individuals as parents. Then, we randomly select a pair of parents each time for crossover and generate a child. Within the crossover process, each element in the child's vector is chosen randomly from one of the parents'. Also, based on the mutation ratio setting, part of the offsprings will further perform mutation operations. For example, with mutation ratio 0.25 and mutation probability 0.1, 250 out of 750 children have a possibility of 0.1 to mutate. If a child is chosen to mutate, its kernel size, expansion ratio, and depth will be randomly sampled out of all the possible values for exploration. After crossover and mutation, we have a new population consisting of parents, bred children, and mutated children. Next, the fittest individuals are selected as new parents for next iteration. The above crossover and mutation steps will be repeated for the maximum evolutionary search iteration number.

### A.2 Evolutionary Search Hyperparameters

Typically, the evolutionary search is not very sensitive against different hyperparameter settings, provided that the population size and iteration number are large enough and that there is adequate exploration. In Section 6, our hyperparameter settings are: population size is 1000, parent ratio is 0.25, mutation probability is 0.1, mutation ratio is 0.25, and we search for 50 generations given each latency constraint. We denote these settings as "EA#1". In Fig. 17, we change the hyperparameters to "EA#2": population size is 500, parent ratio is 0.3, mutation probability is 0.2, mutation ratio is 0.4, and run evolutionary search again on Samsuang Galaxy S5e. The results in Fig. 17 show that the searched Pareto-optimal models are almost identical to the original ones ("EA#1").

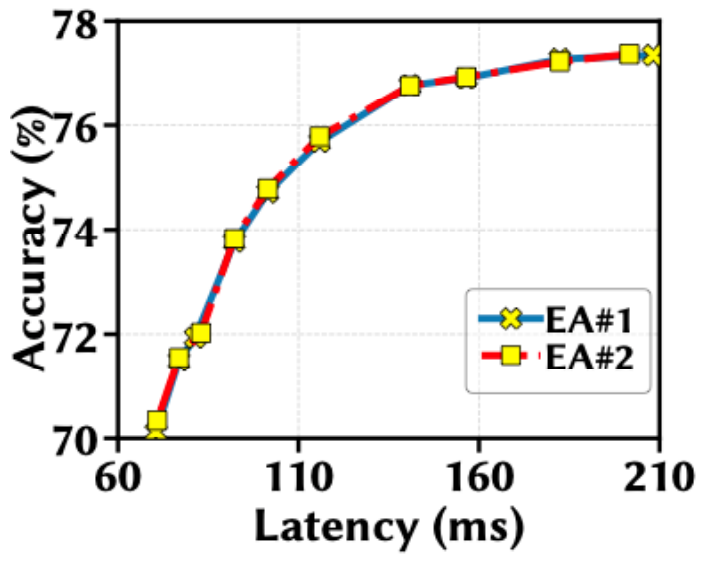

Fig. 17. Pareto-optimal models searched on Samsung Galaxy S5e with different parameter settings for evolutionary search. "EA#1" denotes the parameter setting that population size is 1000, parent ratio is 0.25, mutation probability is 0.1 and mutation ratio is 0.25; while "EA#2" represents that population size is 500, parent ratio is 0.3, mutation probability is 0.2 and mutation ratio is 0.4.

## B ADDITIONAL RESULTS

In this section, we present additional experimental results, including the demonstration of latency monotonicity based on third-party latency results and the effectiveness of our transfer learning technique in various mainstream search spaces.

### B.1 Latency Monotonicity

To corroborate our own measurement and finding in Section 4, we examine latency monotonicity by leveraging third-party latency predictors and measurements for other devices.

*B.1.1 Results on Predicted Latencies.* We obtain latency lookup tables for four mobile devices [7]: Google **Pixel1**, **Pixel2**, Samsung Galaxy **S7** edge, **Note8** in the MobileNet-V2 space with stage widths "32, 16, 24, 48, 80, 104, 192, 320, 1280". In addition, we obtain from [8] latency lookup tables for four cross-platform devices (used in [10]): Google **Pixel1**, **Pixel2**, **TITAN** Xp, **E5**-2640 v4 in the MobileNet-V2 space with different stage widths "32, 16, 24, 40, 80, 96, 192, 320, 1280" and measured with the MKL-DNN library. Note that latency predictors are very accurate (e.g., with an root mean squared error of less than 1% of the average) [5, 10, 15, 47].

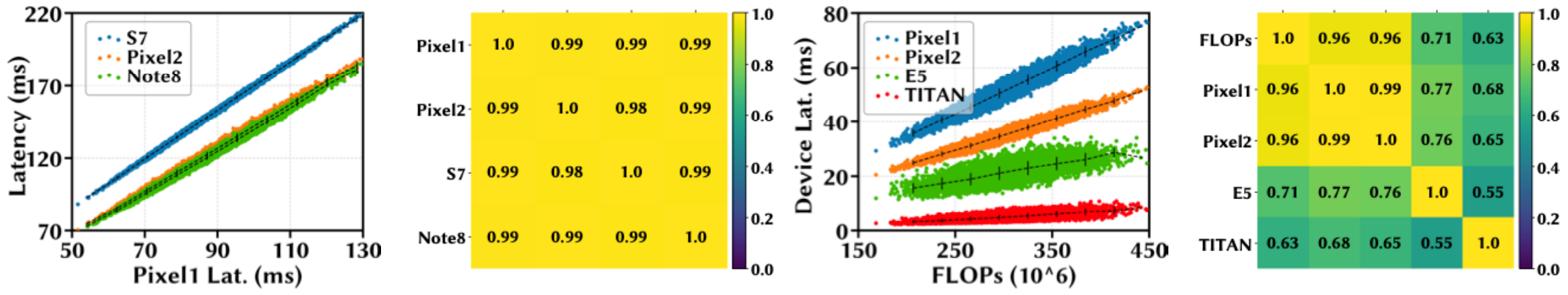

Fig. 18. Latency monotonicity on third-party latency predictors [7, 8]. (a)(b) and (c)(d) use different search spaces and DNN acceleration libraries.

We randomly sample 10k models in each search space with variable depths of "2, 3, 4" in each stage, variable filter sizes of "3, 5, 7" in each convolutional layer, and variable expansion ratios of "3, 4, 6" in each block. We show the results in Fig. 18, which are in line with our experiments: latency monotonicity among mobile devices is strong (>0.95), while FLOP-latency ranking correlation for mobile devices is also quite strong but cross-platform latency monotonicity degrades.

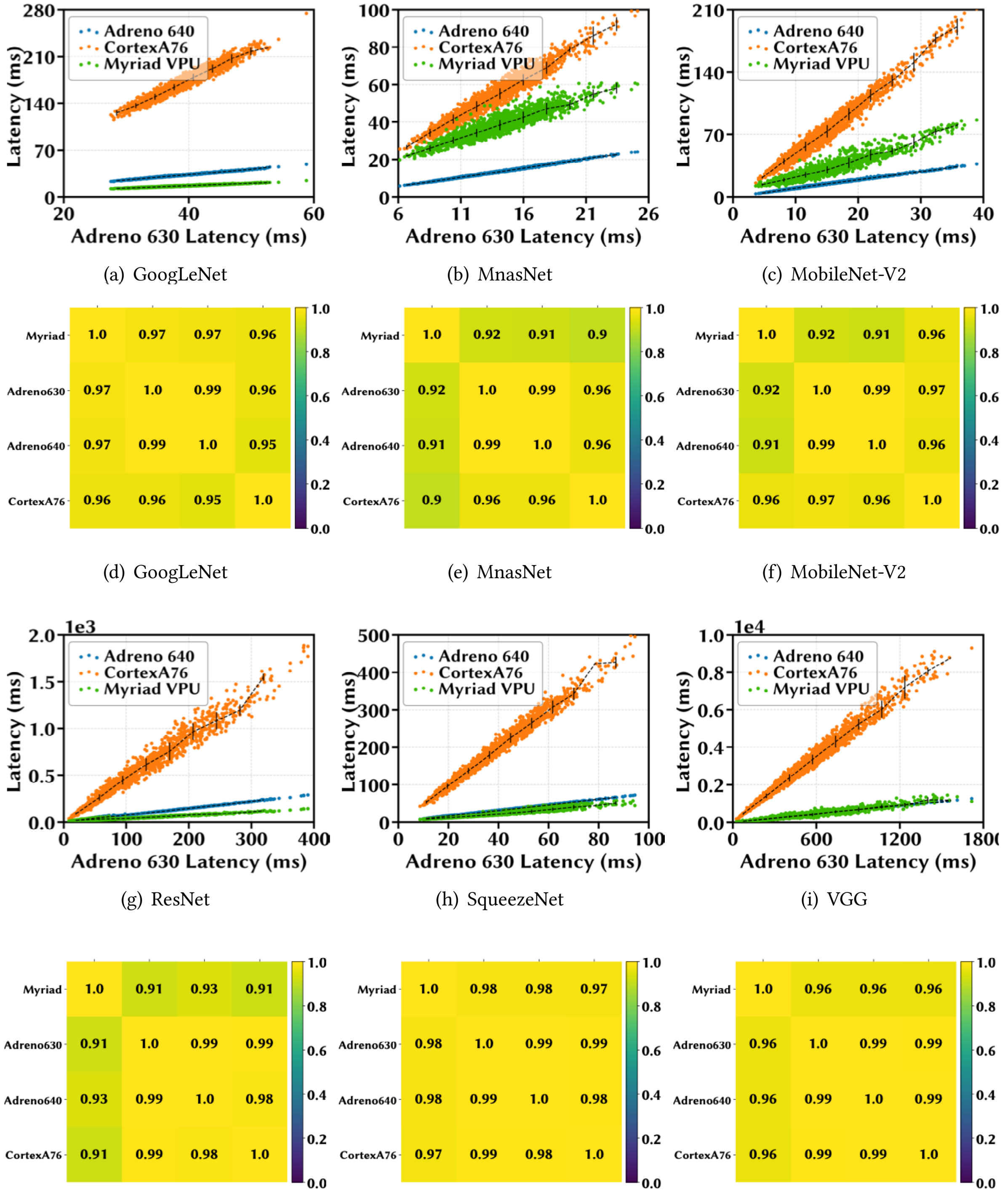

(a) GoogLeNet (b) MnasNet (c) MobileNet-V2

(d) GoogLeNet (e) MnasNet (f) MobileNet-V2

(g) ResNet (h) SqueezeNet (i) VGG

(j) ResNet (k) SqueezeNet (l) VGG

Fig. 19. Latency results of 2000 models on CortexA76 CPU, Adreno 640 GPU, Adreno 630 GPU, and Myriad VPU, available in the dataset [49]. Search spaces: (a)(d) GoogLeNet, (b)(e) MnasNet, (c)(f) MobileNet-V2, (g)(j) ResNet, (h)(k) SqueezeNet, and (i)(l) VGG.

*B.1.2 Results on Measured Latencies.* We provide more evidence of latency monotonicity across different devices, and even across different DNN frameworks, using the nn-Meter results [49, 50]. Specifically, Fig. 19 shows the measured latencies and cross-device SRCCs in six different search spaces. We see that cross-device latency monotonicity strongly exists.

## B.2 Results on MobileNet-V2

**Search Space.** Our backbone is MobileNet-V2 with multiplier 1.3, with the channel number in each block fixed. As shown in Fig. 20, The search space consists of depth of each stage, kernel size of convolutional layers, and expansion ratio of each block. The depth can be chosen from "2, 3, 4", kernel size can be "3, 5, 7", and candidate expansion ratios are "3, 4, 6". There are five stages whose configurations can be searched, plus the kernel size and expansion ratio of the last inverted residual block.

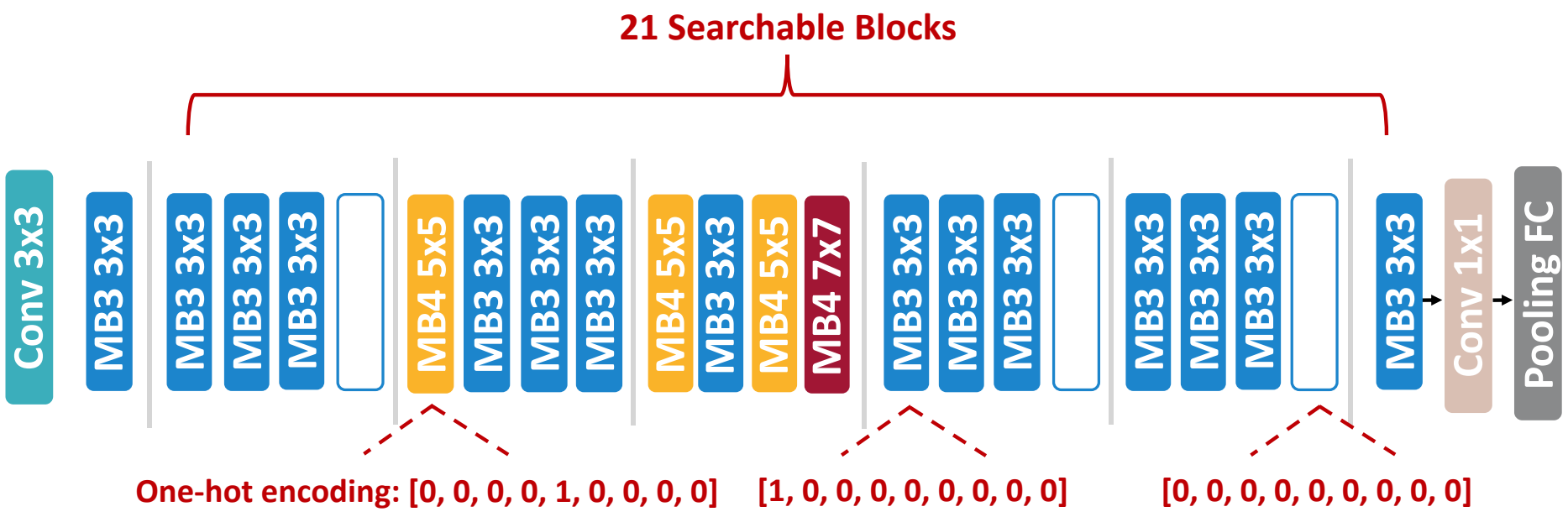

Fig. 20. MobileNet-V2 search space and architectural encoding.

**Proxy Adaptation.** We use S5e as the default proxy device. Fig. 21 shows the original SRCC between S5e and desktop CPUs and GPUs, which are all below 0.8. We observe from Section 5 that SRCC of <0.8 is not enough to find Pareto-optimal architectures on the target device. Thus, in the absence of strong latency monotonicity between the default proxy device and the target device, proxy adaptation is necessary.

In the MobileNet-V2 search space, we have 21 searchable blocks in total, whose configurations can each be chosen out of nine kernel size and expansion ratio combinations or none (i.e., the block is not selected with a reduced stage depth). Thus, to represent an architecture, we simply use a 9-dimension one-hot vector $\mathbf{x}_b$ to encode the specification of each block. Given the proxy device's latency predictor as $L_{\mathbf{d}_0}(\mathbf{x}) = \mathbf{w}^T\mathbf{x}$ built a priori, we collect the latencies of 80 sampled architectures on the target device, which are further split into 60 for training and 20 for validation. For i7-4790 and i7-4770HQ, we only need latencies of 30 sampled architectures for training. Next, by solving Eqn. (3), we obtain the AdaProxy device's latency predictor adapted to the target device, resulting in a significantly increased SRCC. Therefore, with the new latency predictor, we can quickly obtain Pareto-optimal architectures for the AdaProxy device, which are also very close to optimum for the target device (after removal of non-Pareto optimal architectures as specified in Section 5).

**Results.** Even without proxy adaptation, our results in Section 4 show that latency monotonicity among mobile devices (and between S5e and FPGA) is very strong. Here, we show the latency monotonicity between our proxy and desktop GPUs/CPUs, both with and without proxy adaptation. We see from Fig. 21 that the weak monotonicity can be significantly increased by using proxy adaptation. Thus, for a new target device that has a low SRCC with our default proxy device, we can simply use the AdaProxy device's latency predictor instead of profiling thousands of architectures and building a new one.

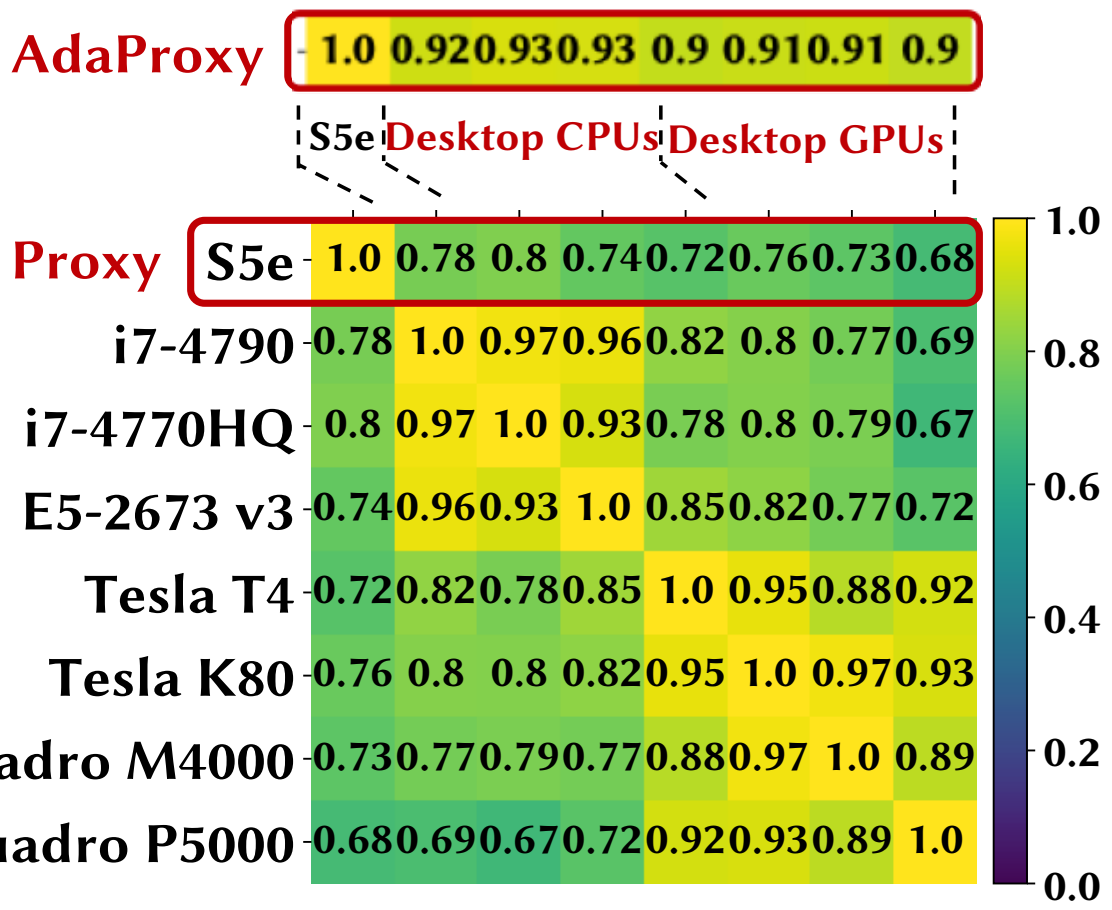

Fig. 21. SRCC for various devices in the MobileNet-V2 space. S5e is the default proxy device. SRCC values boosted by AdaProxy are highlighted.

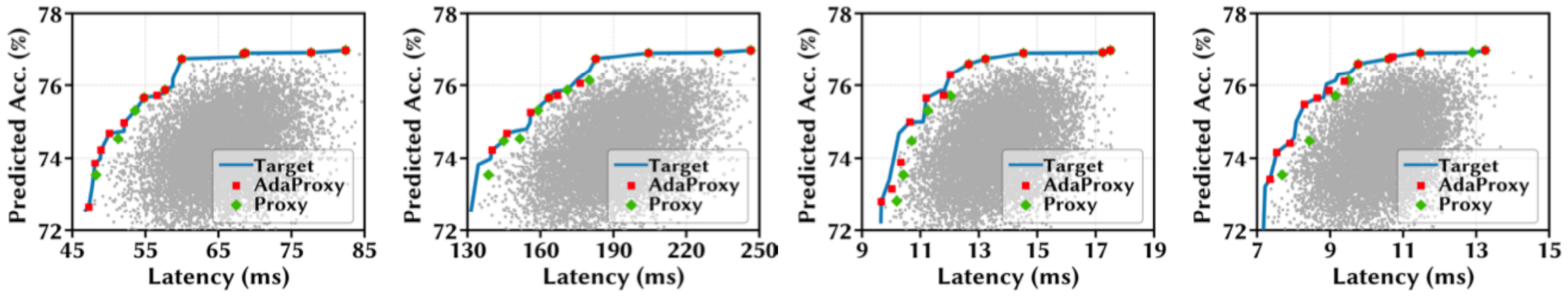

(a) Intel i7-4790 (b) E5-2673 v3 (c) Tesla K80 (d) Quadro P5000

Fig. 22. Exhaustive search results based on 10k random architectures and predicted accuracies, for non-mobile target devices with the default S5e proxy and AdaProxy. SRCC values before and after proxy adaptation are shown in Fig. 21.

In Section 6, we already show the architecture performances for mobile target devices and some GPU/CPU devices. Now, we show the architecture performances for the remaining GPU/CPU devices in Fig. 22. We see that, due to the low SRCC, the architectures found by using the default proxy device's latency predictor may not overlap well with the oracle's Pareto-optimal boundary. On the other hand, with improved SRCC, the architectures found by using the AdaProxy device's latency predictor preserves Pareto optimality very well on the target devices. Again, this shows that our proposed transfer learning approach to boost the latency monotonicity is necessary and effective. For the devices in Fig. 22(d), we use 80 sampled architectures (50 for training, and 30 for validation and tuning $\lambda$) to construct AdaProxy. Note that the results are based on exhaustive search out of 10k randomly selected architectures, using the predicted accuracies as the true values. This is essentially considering a semi-oracle NAS process (on a small space of 10k architectures) assuming a perfect accuracy predictor. In other words, compared to evolutionary search (whose accuracy predictor itself is also not perfect), it has a more stringent requirement on the SRCC between the target device and the proxy (or AdaProxy) device. Thus, our approach works well even in this challenging case.

### B.3 Results on NAS-Bench-201

**Search Space.** NAS-Bench-201 adopts a fixed cell search space [17]. Each searched cell is represented as a densely-connected directed acyclic graph (DAG), which is then stacked together with a pre-defined skeleton to construct an architecture. Specifically, as shown in Fig. 23, the search space considers four nodes and five representative operation candidates for the operation set, and varies the feature map sizes and dimensions of the final fully-connected layer to handle different datasets (i.e., CIFAR-10, CIFAR-100, and ImageNet16-120).

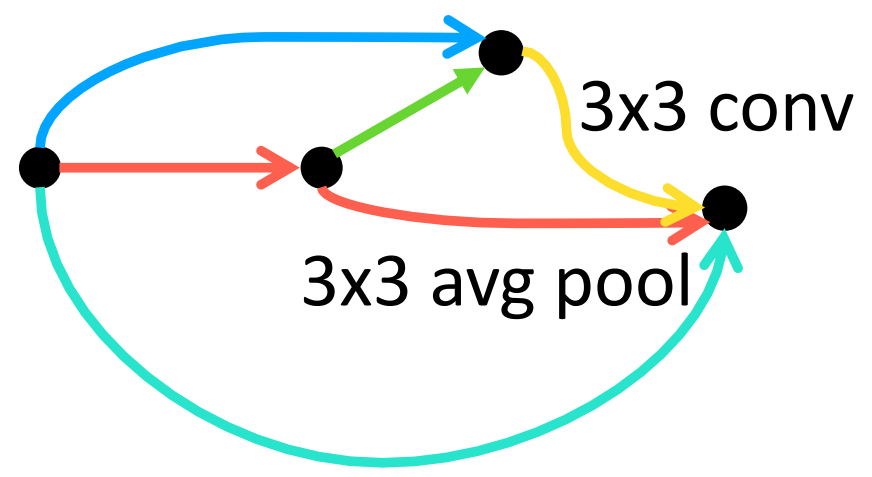

Fig. 23. NAS-Bench-201 search space and architectural encoding.

**Proxy Adaptation.** We have four searchable nodes in total, the operation for each of which can be chosen from five candidates. Thus, we can use a 5-dimension one-hot vector to encode the specification of each node, although more advanced representation (e.g., graph-based [19]) is also applicable. Pixel3 is the default proxy device. Given the proxy device's latency predictor as $L_{\mathbf{d}_0}(\mathbf{x}) = \mathbf{w}^T\mathbf{x}$ built a priori, the training in transfer learning is based on measured latencies of 40 sampled architectures for the edge TPU and edge GPU, and 20 sampled architectures for Eyeriss and FPGA, respectively. In addition, validation uses another 20 sampled architectures for tuning the hyperparameter. Next, by solving Eqn. (3), we obtain the AdaProxy device's latency predictor adapted to the target device, resulting in a significantly increased SRCC. We show in Fig. 24 the latency monotonicity in terms of SRCC values, both with and without proxy adaptation. We see that the weak monotonicity can be significantly increased by using proxy adaptation.

**Results.** Considering the CIFAR-100 dataset, Fig. 25 shows optimal architectures found by using the proxy device's latency predictor, the adapted latency predictor, and the oracle, respectively. We can see that due to the pre-adaptation low SRCC values between the proxy device Pixel3 and the target devices, only a few architectures that are optimal for the proxy are still optimal for the target devices after architecture removal. Moreover, even the proxy's remaining optimal architectures can be far from optimality on the target device. For example, Fig. 25(a) shows that some of Pixel3's optimal architectures deviate from the Pareto-optimal boundary on the edge GPU. By using proxy adaptation and increasing the SRCC values, the AdaProxy's optimal architectures can be efficiently transferred to target devices while preserving optimality. The proxy device Pixel3 has a high SRCC of 0.96 with Raspi4, even without proxy adaptation. Thus, as shown in Fig. 25(e), the optimality of Pixel3's architectures preserve very well on Raspi4. All these demonstrate the importance of strong monotonicity between the proxy and the target device, as well as the effectiveness of our proxy adaptation technique, for hardware-aware NAS with a total latency evaluation cost of $O(1)$.

The same observation is also made in Fig. 26 for the ImageNet16-120 dataset.

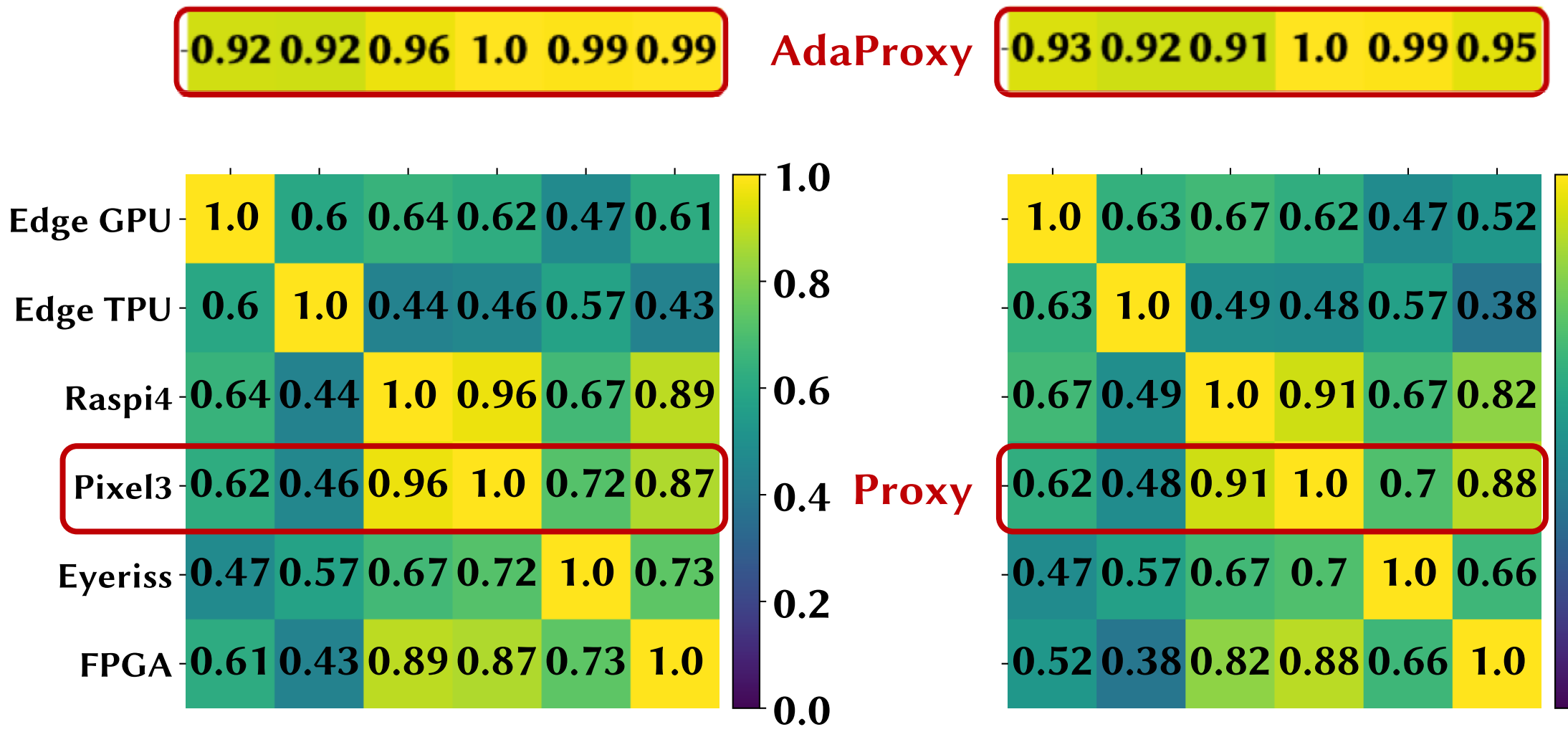

Fig. 24. SRCC for various devices in the NAS-Bench-201 search space on CIFAR-100 (left) and ImageNet16-120 (right) datasets. Pixel3 is our proxy device. SRCC values boosted with AdaProxy are highlighted.

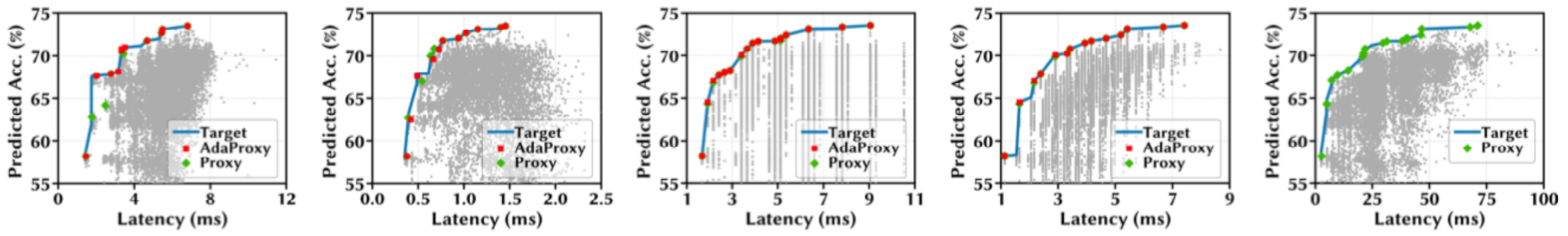

(a) Edge GPU (b) Edge TPU (c) Eyeriss (d) FPGA (e) Raspberry Pi 4

Fig. 25. Exhaustive search results for different target devices on NAS-Bench-201 architectures (CIFAR-100 dataset) [17, 29]. Pixel3 is the proxy. SRCC values before and after proxy adaptation are shown in the left subfigure of Fig. 24.

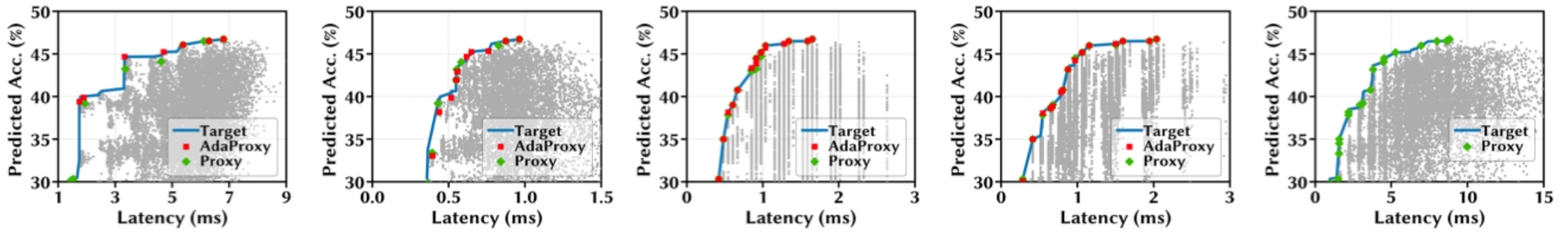

(a) Edge GPU (b) Edge TPU (c) Eyeriss (d) FPGA (e) Raspberry Pi 4

Fig. 26. Exhaustive search results for different target devices on NAS-Bench-201 architectures (ImageNet16-120 dataset) [17, 29]. Pixel3 is the proxy. SRCC values before and after proxy adaptation are shown in the right subfigure of Fig. 24.

## B.4 Results on FBNet

**Search Space.** Similar to MobileNet-V2, the FBNet search space is also layer-wise with a fixed macro-architecture, which defines the number of layers and input/output dimensions of each layer and fixes the first and last three layers, with the remaining layers to be searched. As shown in Fig. 27, the overall search space consists of 22 searchable blocks: the first and last inverted residual blocks, and five stages within each of which there are at most four blocks. For each block, the kernel size can be chosen from "3, 5", and the expansion ratio can be "1, 3, 6". For the first and last 1x1 convolution layer, *group* convolution can be used to reduce the computation complexity. Also, each block can be skipped. Thus, there are nine candidate specification choice for each block (detailed configurations are shown in Table 2 of [45]).

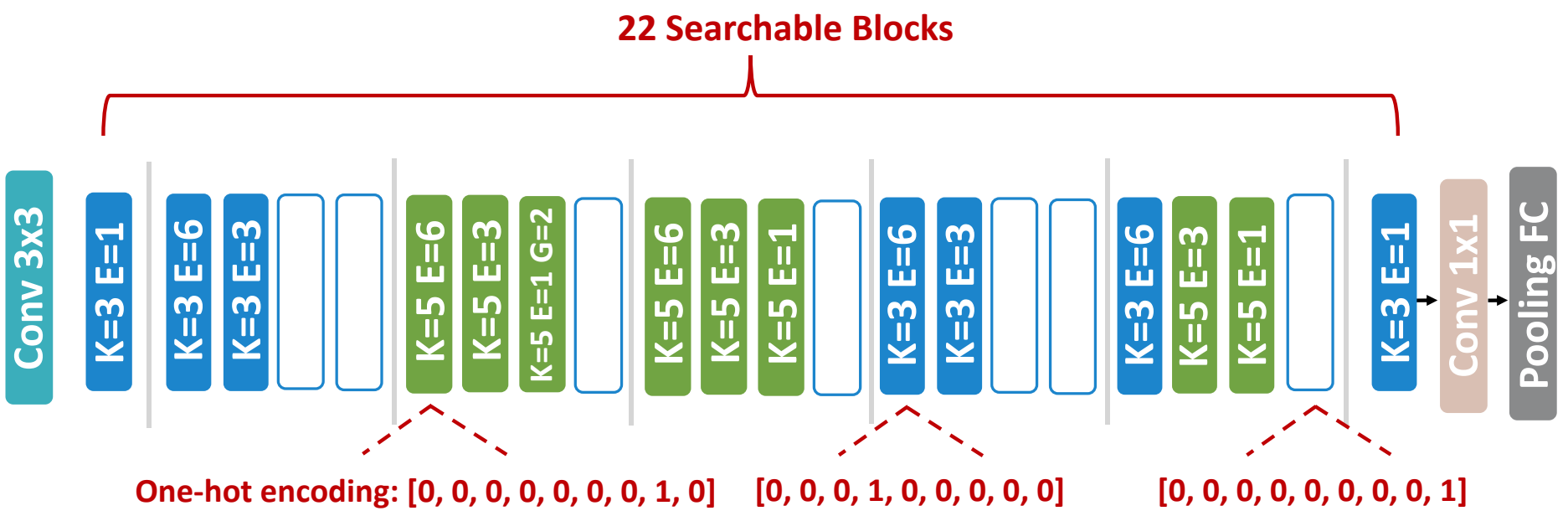

Fig. 27. FBNet search space and architectural encoding.

**Proxy Adaptation.** We have 22 searchable blocks in total, the configuration for each of which can be chosen from the nine architecture candidates (including "Skip"). Then, we can still use a 9-dimension one-hot vector to encode each block. Using Pixel3 as the default proxy and the same approach as in Appendix B.2, we can solve Eqn. (3) to create an AdaProxy device, which has SRCC of close to 0.9 or higher with the target device. In the transfer learning process, the numbers of sampled architectures for training are: 80 (Edge GPU), 40 (Raspi4), 30 (FPGA), 20 (Eyeriss). In addition, validation uses another 20 sampled architectures for tuning the hyperparameter.

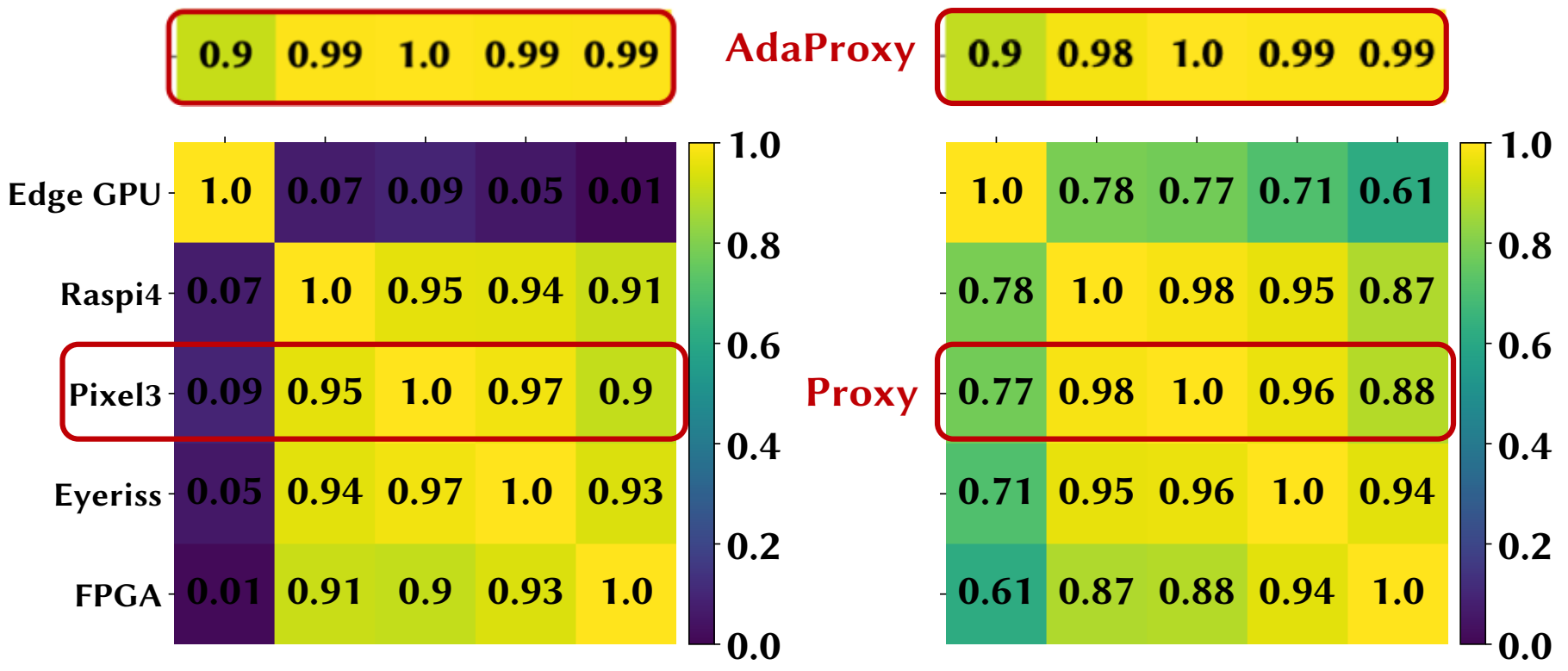

Fig. 28. SRCC for various devices in the FBNet search spaces [29], on CIFAR-100 (left) and ImageNet16-120 (right) datasets respectively. Pixel3 is the proxy. SRCC values boosted by AdaProxy are highlighted.

**Results.** Our key focus is to achieve a high SRCC between the proxy (or AdaProxy) device and the target device, such that we can efficiently transfer the optimal architectures found on the proxy (or AdaProxy) device to the new target device without measuring latencies of thousands of architectures and building a new latency predictor. Since the accuracy results for architectures in the FBNet search space are not available [29], we only show in Fig. 28 the SRCC values instead, both with and without proxy adaptation. We can see that cross-platform SRCCs are greatly boosted (i.e., close to 1) with AdaProxy. By Theorem 3.1, the strong latency monotonicity ensures that the optimal architectures found on the proxy (or AdaProxy) device can be applied to new target devices.

## B.5 Results on nn-Meter

The nn-Meter dataset released in [49, 50] includes measured inference latencies of 2000 models from 11 search spaces, including GoogLeNet, MnasNet and ProxylessNAS, etc on three mobile devices and one edge device: Pixel4 (Cortex A76 CPU), Mi9 (Adreno 640 GPU), Pixel3XL (Adreno 630 GPU), and Myriad VPU (Intel Movidius NCS2 edge device). Fig. 19 shows that the devices already have strong latency monotonicity with SRCC values greater than 0.9 on six search spaces. Among the remaining five search spaces, MobileNet-V1 and AlexNet are obsolete and phased out for hardware-ware NAS. Next, we apply our proxy adaptation technique on the other three search spaces: **MobileNet-V3**, **NAS-Bench-201**, and **ProxylessNAS**, which are mainstream and widely-used backbones in SOTA NAS algorithms.

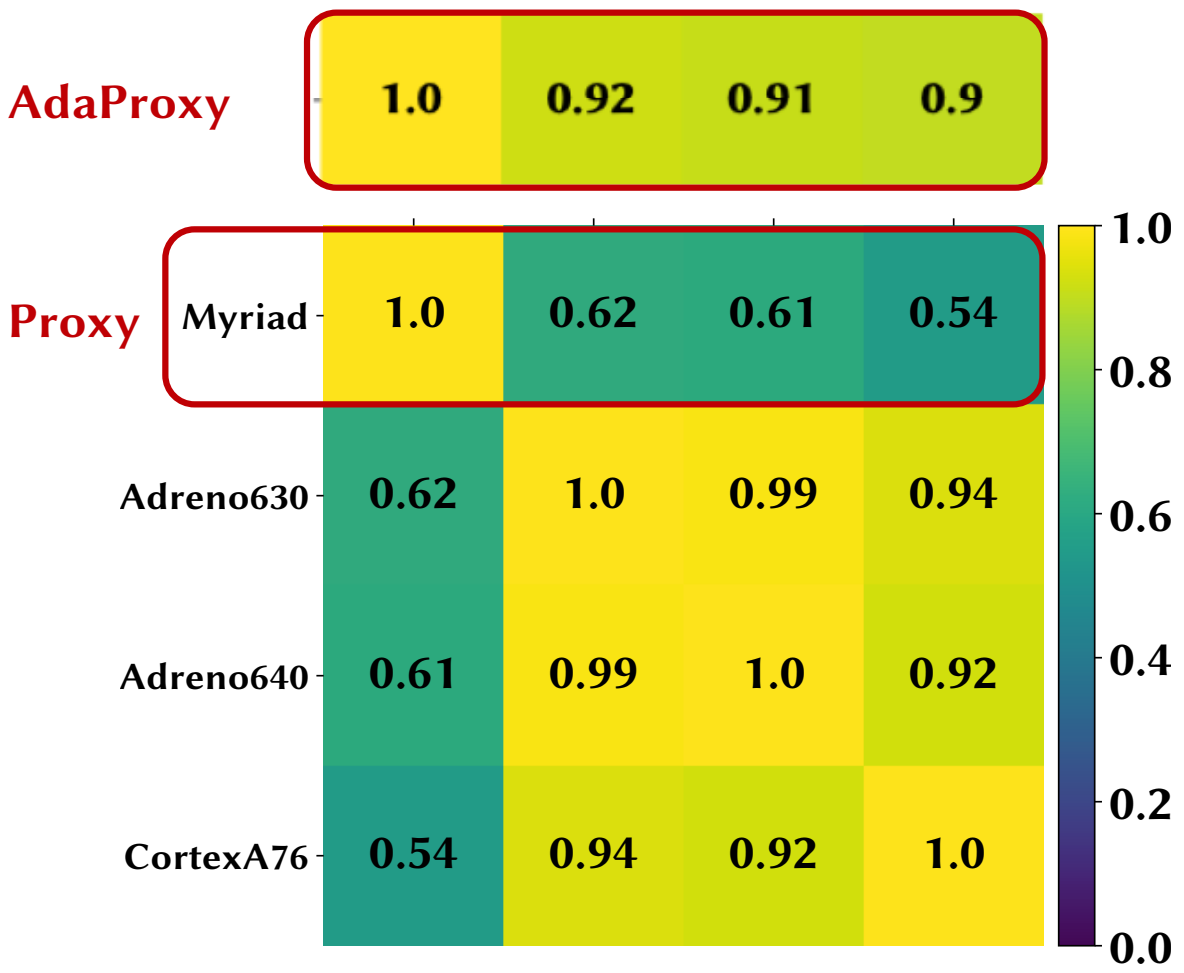

Fig. 29. SRCC for various devices in the MobileNet-V3 search space [49, 50]. SRCC values boosted by AdaProxy are highlighted.

*B.5.1 MobileNet-V3.* In our experiment, the number of searchable blocks in the MobileNet-V3 space is fixed as 12. For each block, the input, mid, and output channel number, and kernel size are variable from a set of candidates. Instead of directly using the kernel-based latency predictor in [50] that has a very large dimensionality for one-hot architectural encoding, we use a simple block-level encoding method. Concretely, for each block, we use one-hot encodings for the input, mid, and output channel number and kernel size, respectively, and then concatenate these four one-hot vectors together to get the block-level encoding. After further concatenating the encoding vector of each block, we have a 530-dimension encoding for each architecture. Then, we build a simple 4-layer fully-connected neural network (with 500/250/100 neurons in each hidden layer) and train it on the latency data of the edge device Myriad VPU (used as the proxy), which has a low SRCC with the other three mobile devices. For the neural network training, we split the 1000 data samples (we use 1000 out of 2000 models for this experiment) into 800 for training and 200 for testing, set the learning rate as 0.01 and the batch size as 128, and train the network for 500 epoches. We also compress the network to 2 layers by fixing the first layer and appending it with another layer for the proxy device's latency predictor. Next, we apply the transfer learning method in Section 5.4 to the three target mobile devices. We use latencies of 150 architectures for transfer learning on Adreno 630/640 and 160 architectures for Cortex A76, respectively, while using 20 architectures for validation. The relatively larger number of latency measurements needed for boosting the latency monotonicity is due in great part to two reasons: (1) MobileNet-V3 is a fairly complex search space, with many searchable operators; and (2) we intentionally address a challenging case where the proxy device has weak monotonicity with all the target devices. The results are shown in Fig. 29, where we can see that the SRCC values are significantly increased after proxy adaptation. despite the initially weak latency monotonicity.

*B.5.2 ProxylessNAS.* This search space is based on the MobileNet-V2 backbone, with variable expansion ratios, kernel sizes, inputs, and output channel numbers [10]. We apply a similar encoding approach as in the MobileNet-V3 space, and get a 783-dimension vector for each architecture in the nn-Meter dataset [49]. The Myriad VPU and Adreno 640 GPU is the only pair of devices with SRCC less than 0.9, with the pre-adaptation SRCC already being 0.87. We directly use the 783-dimension vector to perform transfer learning by updating the weights pre-trained on the proxy device (Adreno 640 GPU), with latencies of 30 sampled architectures for training and 20 architectures for validation. The results are shown in Fig. 30, demonstrating that the SRCC can be increased to over 0.9 after proxy adaptation.

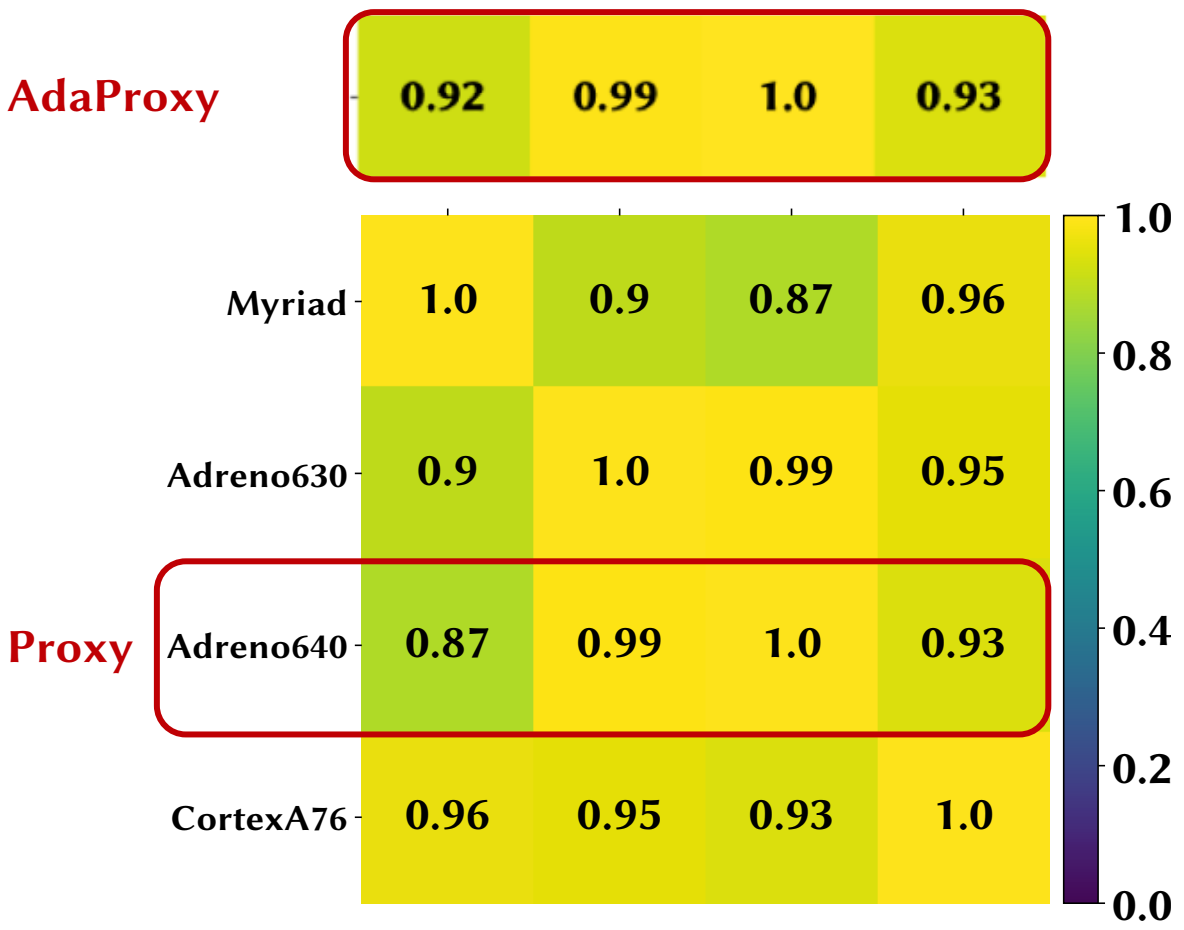

Fig. 30. SRCC for various devices in the ProxylessNAS search space [49, 50]. SRCC values boosted by AdaProxy are highlighted. We only apply proxy adaptation for the Myriad VPU edge device, since the other target devices already have high SRCC of 0.9+ with the proxy device.

*B.5.3 NAS-Bench-201.* For the NAS-Bench-201 space, we adopt the same encoding method as described in Appendix B.3. We also consolidate the latency datasets released by three different research studies [19, 29, 50] for the NAS-Bench-201 search space. The Myriad VPU edge device is the default proxy, while the target devices include FPGA, GPU, CPU, mobile, edge device, DSP, and TPU. Using the latencies of 20 sampled architectures for validation, the numbers of sampled architectures for training in the transfer learning process are: 30 for Edge GPU, Edge TPU, Eyeriss, FPGA, Raspi4, Adreno 630, Adreno 640, Cortex A76, CPU 855, GPU 855, 50 for DSP 855, 55 for Pixel3 and Jetson, 60 for GTX and i7, and 90 for Jetson 16. Note that the dataset in [49] only contains latencies for 2000 architectures in the NAS-Bench-201 space, and hence we only consider these 2000 architectures when calculating the cross-device SRCC values. We show the results in Fig. 31. While the latencies are measured by different research groups, on very different devices and using different deep learning frameworks, our proxy adaptation technique can still successfully increase the SRCC values to 0.9+, significantly boosting the otherwise weak latency monotonicity and keeping the total latency evaluation cost at $O(1)$ for hardware-aware NAS.

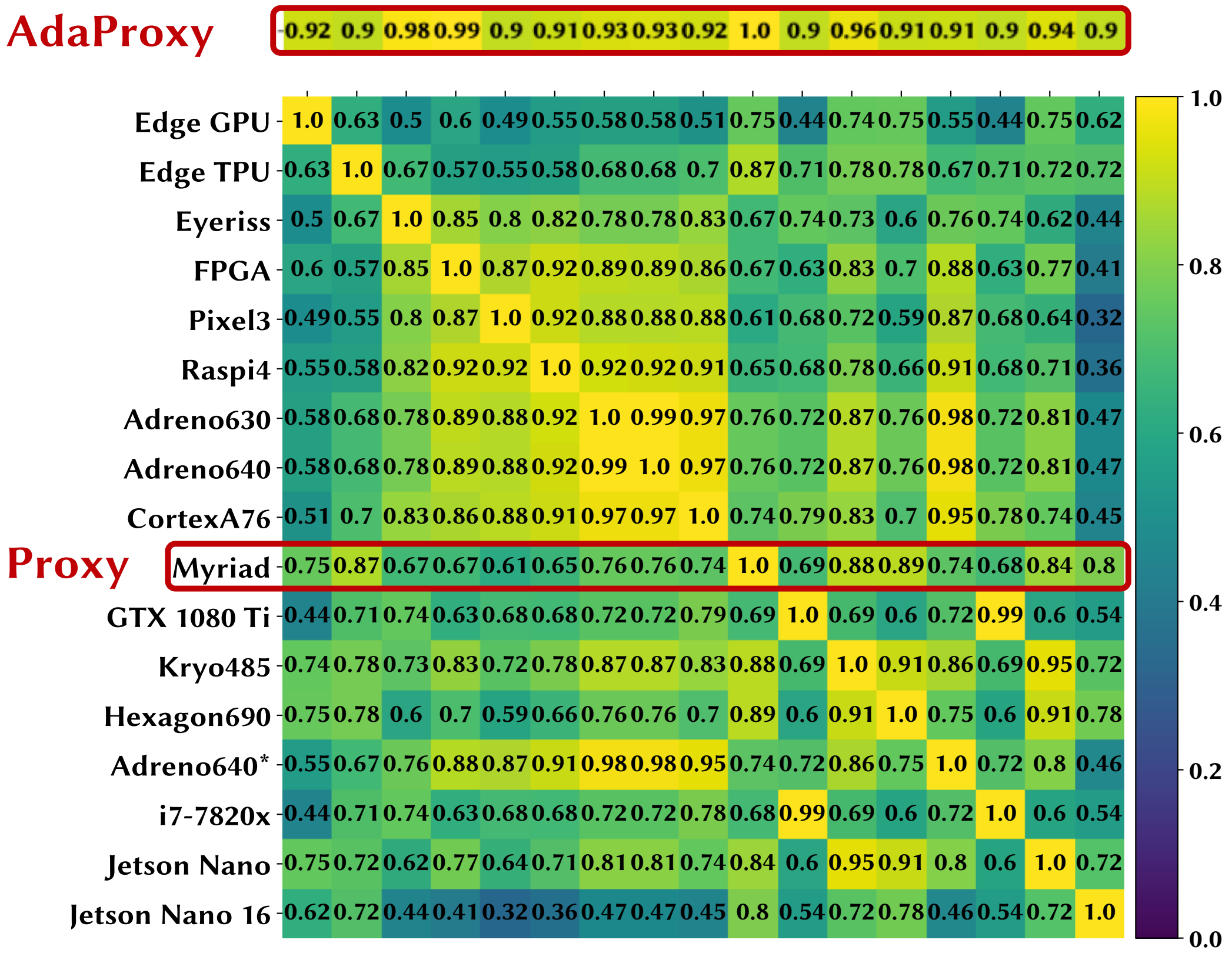

Fig. 31. SRCC for various devices in the NAS-Bench-201 search space with latencies collected from [19, 29, 49, 50]. SRCC values boosted by AdaProxy are highlighted. "Adreno640" and "Adreno640*" denote model latencies measured by [50] and [19] respectively. "Jetson Nano" and "Jetson Nano 16" represent the latencies of FP32 and FP16 models correspondingly.